\documentclass[10pt,letterpaper]{article}

\pdfoutput = 1

\usepackage[top=0.85in,left=2.75in,footskip=0.75in]{geometry}

\usepackage{changepage}

\usepackage[utf8]{inputenc}

\usepackage{textcomp,marvosym}

\usepackage{fixltx2e}

\usepackage{amsmath,amssymb}

\usepackage{cite}

\usepackage{nameref,hyperref}

\usepackage[right]{lineno}

\usepackage{microtype}
\DisableLigatures[f]{encoding = *, family = * }

\usepackage{rotating}

\raggedright
\usepackage[aboveskip=1pt,labelfont=bf,labelsep=period,justification=raggedright,singlelinecheck=off]{caption}

\makeatletter
\renewcommand{\@biblabel}[1]{\quad#1.}
\makeatother

\date{}

\usepackage{lastpage,fancyhdr,graphicx}
\usepackage{epstopdf}
\fancyheadoffset[L]{2.25in}
\fancyfootoffset[L]{2.25in}

\usepackage{enumerate}
\usepackage{booktabs}
\usepackage{multirow}
\usepackage{algorithm}
\usepackage{algorithmicx}
\usepackage{algpseudocode}
\usepackage{subfigure}

\begin{document}
\vspace*{0.35in}

\begin{flushleft}
{\Large
\textbf\newline{Modern Physiognomy: An Investigation on Predicting Personality Traits and Intelligence from the Human Face}
}
\newline
\\
Rizhen Qin\textsuperscript{1},
Wei Gao\textsuperscript{1},
Huarong Xu\textsuperscript{3},
Zhanyi Hu\textsuperscript{1,2*}
\\
\bigskip
\bf{1} National Laboratory of Pattern Recognition, Institute of Automation, Chinese Academy of Sciences, Beijing, 100190, China
\bf{2} CAS Center for Excellence in Brain Science and Intelligent Technology, Institute of Automation, Chinese Academy of Sciences, Beijing, 100190, China
\bf{3} Department of Computer Science and Technology, Xiamen University of Technology, Xiamen, Fujian Province, 361024, China
\\
\bigskip

%
%





* huzy@nlpr.ia.ac.cn

\end{flushleft}
\section*{Abstract}
The human behavior of evaluating other individuals with respect to their personality traits and intelligence by evaluating their faces plays a crucial role in human relations. These trait judgments might influence important social outcomes in our lives such as elections and court sentences. Previous studies have reported that human can make valid inferences for at least four personality traits. In addition, some studies have demonstrated that facial trait evaluation can be learned using machine learning methods accurately. In this work, we experimentally explore whether self-reported personality traits and intelligence can be predicted reliably from a facial image. More specifically, the prediction problem is separately cast in two parts: a classification task and a regression task. A facial structural feature is constructed from the relations among facial salient points, and an appearance feature is built by five texture descriptors. In addition, a minutia-based fingerprint feature from a fingerprint image is also explored. The classification results show that the personality traits ``Rule-consciousness'' and ``Vigilance'' can be predicted reliably, and that the traits of females can be predicted more accurately than those of male. However, the regression experiments show that it is difficult to predict scores for individual personality traits and intelligence. The residual plots and the correlation results indicate no evident linear correlation between the measured scores and the predicted scores. Both the classification and the regression results reveal that ``Rule-consciousness'' and ``Tension'' can be reliably predicted from the facial features, while ``Social boldness'' gets the worst prediction results. However, for the median values of the traits, the fitting performance looks better. Finally the classification and regression results show that it is difficult, if not impossible, to predict intelligence from either the facial features or the fingerprint feature, a finding that is in agreement with previous studies.


\section*{Introduction}
Even as long ago as the ancient Chinese, Egyptian and Greek civilizations, people had tried to establish a relationship between facial morphological features and an individual's personality traits~\cite{bib1}. Modern psychological studies have revealed that people tend to evaluate others on their appearance and then proceed to interact with them based on these first impressions. Currently, it has been well established that faces play a central role in people's everyday assessments of other people~\cite{bib2}.

Automatic facial evaluation is a human mechanism that evolved through the process of assessing facial cues. There is the truth behind first impressions: it was shown that humans can make valid inferences for at least four personality traits (Agreeableness, Conscientiousness, Extraversion and Dominance) from facial features~\cite{bib3,bib4}. In~\cite{bib5}, Wolffhechel et al. studied the relationship between self-reported personality traits and first impressions and found that some personality traits could be predicted from faces to a certain extent. The results of~\cite{bib6} showed that the people could accurately evaluate the intelligence of males by viewing their faces.

To simulate human communicative behaviors,~\cite{bib7,bib8} used machine learning methods to construct an automatic trait predictor based on facial structural descriptors and appearance descriptors. They found that all the analyzed personality traits could be predicted accurately. As the behaviors used by humans to evaluate personality traits contain some commonalities, these behaviors can also be simulated by machine learning methods.

Currently, the following two points hold: First, some self-reported personality traits and intelligence can be evaluated to a certain extent by humans based on facial features. Second, the commonalities existing in the evaluation behavior of the human can be mined by machine learning methods. In~\cite{bib5,bib6}, the relationships between the facial features and self-reported personality traits or measured intelligence were studied. However, their findings were mostly negative; little correlation was found between facial features and personality traits or measured intelligence.

In this work, we further explore whether self-reported personality traits and measured intelligence can be predicted from the facial features by gathering more samples, extracting more diverse features, and adopting more classification and regression rules. To represent the characteristics of the facial images, we construct a structural feature according to the description in~\cite{bib7} and construct an appearance feature using five textural descriptors: HOG~\cite{bib10}, LBP~\cite{bib11}, Gabor~\cite{bib12}, Gist~\cite{bib13} and SIFT~\cite{bib14}. For these extracted features to be more informative for facial representation, the facial images are preprocessed by segmentation and cropping to remove irrelevant information such as the background, hair and clothes. In addition, an image pyramid is built for each face image to form a multi-scale high-dimensional representation. In our work, we also perform experiments to study the relationships between the fingerprint feature and self-reported personality traits as well as measured intelligence to see whether we could obtain consistent results for both the facial image and the fingerprint image.

The above three types of features are used as the sample representations for the personality traits prediction in this work, and the prediction problem is cast separately as both a classification task and a regression task, where a bank of classification and regression methods are systematically investigated. To construct the classification labels and the regression targets, the personality traits scores and the measured intelligence scores are converted into appropriate values first, and then, a variety of criteria are used to evaluate the experimental results.

Our experiments show that some personality traits are related to the tested facial characteristics and can be predicted from those facial features fairly reliably, while some other personality traits may largely depend on the social environment; little correlation seems to exist between them and facial characteristics. We find the predictability of the personality traits tends to be more reliable based on the appearance feature rather than on the structural feature or the fingerprint feature. As for the measured intelligence, neither the facial features nor the fingerprint feature provide any reliable prediction for either males or females. In our work, the results of the regression experiments show no evident linear correlation between the predicted scores and the measured scores of the personality traits and intelligence, so it is difficult, if not impossible, to predict their precise scores. However, the prediction performance of the median scores of the personality traits and the high scores of the intelligence is relatively better.

\section*{Related Work}
In~\cite{bib1}, McNeill reported that humans have tried to establish relationships between facial morphological features and individual personality traits since very ancient civilizations such as those in ancient China, Egypt and Greece. At present, psychological studies show that faces play an important role in people's everyday assessments of other people~\cite{bib2}. Humans perform trait judgements from faces unconsciously, and this unconscious behavior can sometimes decisively affect the results of important social events such as elections~\cite{bib15,bib16} or court sentences~\cite{bib17}.

In~\cite{bib18}, Oostehof and Todorov developed a 2D model to identify the basic underlying dimensions of human facial traits evaluation. The authors used the Principal Component Analysis (PCA) technique on the linguistic judgement of the traits and identified two fundamental dimensions: Valence and Dominance. The authors of~\cite{bib19,bib20} studied the human tendency to evaluate others on their faces and identified some important facial features that generate first impressions. In addition, they showed that humans can make valid inferences for at least four personality traits (Agreeableness, Conscientiousness, Extraversion, and Dominance) from facial features.

Automatic evaluation of faces is a human mechanism that evolved from the need to quickly assess potential threats-an evolution that has shaped the saliency of facial cues~\cite{bib7}. In~\cite{bib5}, Wolffhechel et al. studied the relationship between self-reported personality traits and first impressions. Their results revealed that some personality traits could be inferred from faces to a certain extent. In addition, they found that, on average, people assess a given face in a highly similar manner. In~\cite{bib6}, the authors used facial photographs of 40 men and 40 women to test the relationship between measured IQ and perceived intelligence. They found that people were able to accurately evaluate the intelligence of men by viewing their facial photographs, but not the intelligence of women.

As stated above, humans can accurately evaluate the personality traits and intelligence of other people to some extent by viewing their faces; therefore, some researchers have investigated whether the trait evaluations performed by humans can be learned automatically by computers. In~\cite{bib7}, Rojas et al. designed a computational system and used the geometrical information contained in a small number of facial points to train the model. Their results suggested that facial traits evaluation can be learned by machine learning methods. In~\cite{bib8}, Rojas et al. carried out further experiments to determine whether facial appearance or facial structural information is more useful for facial traits evaluation. They used a classification framework to evaluate both the structural and appearance information and found that the appearance information was more strongly related to the prediction capability.

Recently, some researchers have investigated whether self-reported personality traits and measured intelligence could be evaluated accurately based on facial characteristics. In~\cite{bib5}, Wolffhechel et al. used a normative self-reported questionnaire (Cubiks In-depth Personality Questionnaire, CIPQ 2.0) to measure participants' personality traits. At the same time, they applied an appearance model to extract both textural and shape information inside the facial boundary. The results (using various nonlinear approaches) revealed that the correlation was not strong enough for a stable prediction. In~\cite{bib6}, the authors tested the relationship between measured intelligence and facial shape, but could establish no correlation between facial morphological features and intelligence as measured with an IQ test.

In this work, we further investigate the predictability of personality traits from facial images by gathering more samples, extracting more diverse features, and adopting more classification and regression rules. The subsequent sections will elaborate on the 4 key components of our work: data acquisition, feature extraction, classification, and regression.

\section*{Materials}
\subsection*{Ethics Statement}
The Institutional Review Board of the Institute of Automation of the Chinese Academy of Sciences has approved this research. The participants were asked to give verbal informed consent to participate in the research and no data were collected until this consent was obtained. The consent is thereby documented by the recording of the data. This was in accordance with the guidelines of the Ethics Committee, Ministry of Health of the People's Republic of China which state that written consent is only required if biological samples are collected, which was not the case in this study. In addition, the data for the measured intelligence, the self-reported personality traits and the fingerprint images were analyzed anonymously.

\subsection*{Photographs}
Facial photographs of 186 students (94 men and 92 women) from the College of Computer And Information Engineering and School of Business, Xiamen University of Technology, China, were used as stimuli. The participants were asked to sit in front of a white background and were photographed with a digital camera, a Canon 5D Mark \uppercase\expandafter{\romannumeral3}. The participants were requested to show a neutral, non-smiling expression and avoid facial cosmetics, jewelry, and other decorations. In addition, we also used a fingerprint sampler to collect fingerprint images from the participants' thumbs.

\subsection*{Personality Measurements}
\begin{figure}[h]
  \centering
  {\includegraphics[width=13.5cm,height=7.0cm]{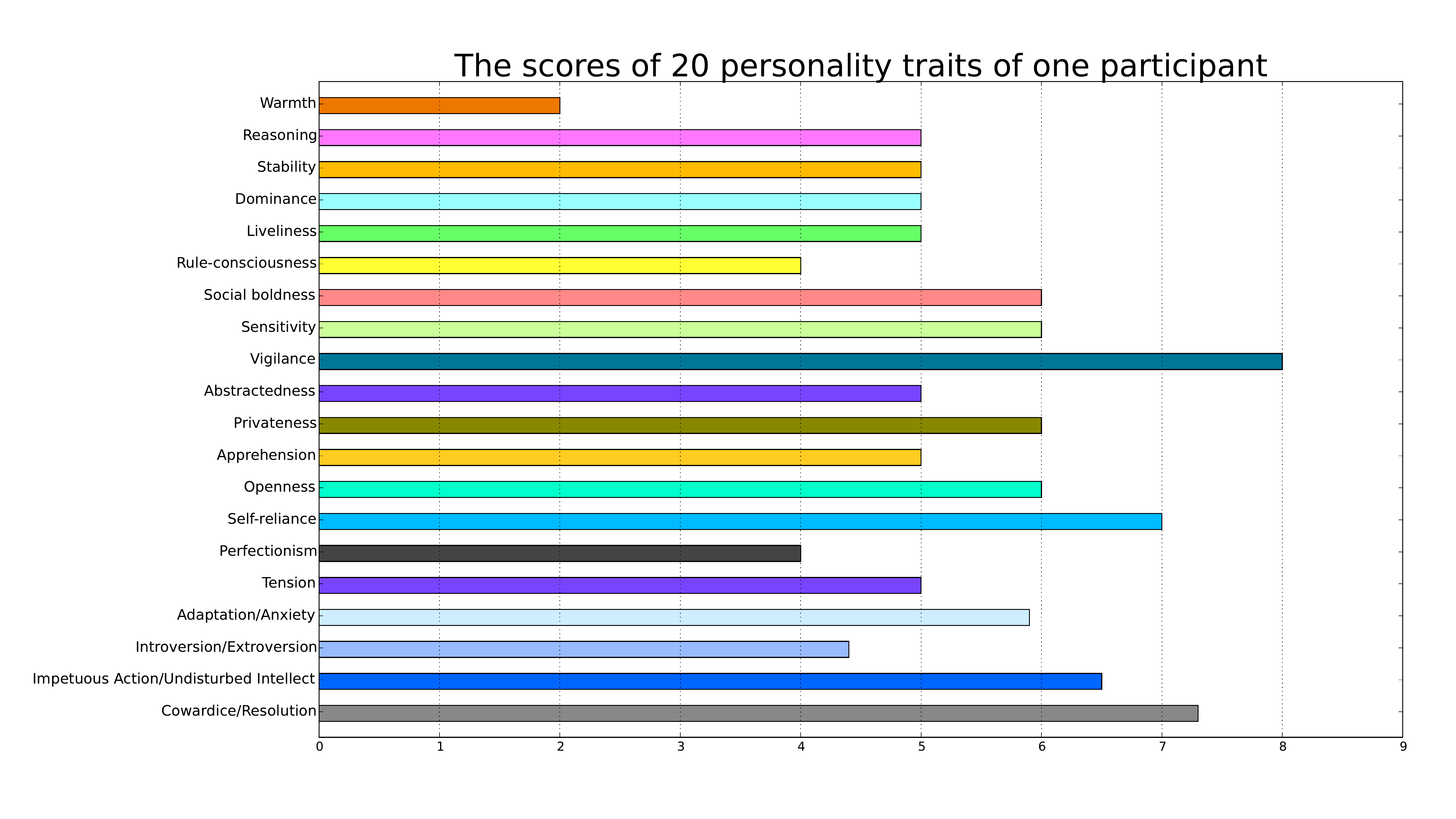}}
\caption{The scores of 20 personality traits for one participant}
\label{fig:1}
\end{figure}

To measure the personality traits of the participants, each was instructed to fill out a Cattell Sixteen Personality Factors Questionnaire (16PF)~\cite{bib21} which is a normative self-report questionnaire that scores 16 personality traits. This questionnaire was compiled by Prof. Cattell, who was engaged in Personality Psychology research for many years. In contrast to other similar tests, the 16PF can capture more personality traits at the same time. Completing the questionnaire required approximately 40 minutes. The traits measured by 16PF are Warmth, Reasoning, Emotional stability, Dominance, Liveliness, Rule-consciousness, Social boldness, Sensitivity, Vigilance, Abstractedness, Privateness, Apprehension, Openness to change, Self-reliance, Perfectionism and Tension. For each personality trait, a score ranging from 1 to 10 was assessed by the commercial software developed by Beijing Normal University Education Training Center according to the responses of each participant on the questionnaire. Note that Prof. Cattell performed a second-order factor analysis based on these sixteen personality factors and acquired the following four second-order factors: Adaptation/Anxiety, Introversion/Extroversion, Impetuous Action/Undisturbed Intellect, and Cowardice/Resolution. In our experiments, we use these 20 traits to describe the personality of each participant. Fig.~\ref{fig:1} shows an example of the 20 personality trait scores for one participant.

\subsection*{Intelligence Measurements}
The Raven's Standard Progressive Matrices (SPM)~\cite{bib22}, which was created by the British Psychologist Raven in 1938, was used to measure participants' intelligence. This test primarily measures the participant's observational ability and ability to think clearly. There is no time limit for the test, but participants finished it in approximately 40 minutes. The total score for the right answers is calculated and converted to a percentile score to measure the intelligence level of the participant. This intelligence metric comprises 60 questions divided into five groups, A, B, C, D and E, with 12 questions in each group. The problem difficulty in the five groups increases gradually and the internal problems within each group are also arranged sequentially by difficulty, from easy to difficult. The thought process required to complete the questions in each group is different. In our experiments, 186 participants completed the test and we use the percentile scores to indicate their intelligence levels.

\section*{Structural, Appearance and Fingerprint Features}
To explore whether the self-reported personality traits and measured intelligence can be evaluated based on facial characteristics, we first extracted the discriminative information contained in the facial images (the photographs taken) of the participants in our experiments. In previous studies, various methods were employed to study the differences and commonalities among faces, including the pixel values, 3-D dimensional scans of faces, annotations of facial landmarks, and so on. In the experiments of ~\cite{bib7,bib8}, the authors employed structural and appearance features of the face, respectively. In our work, we extract the facial characteristics of the samples similarly to~\cite{bib7,bib8}. In the following subsections, the extraction of structural, appearance and fingerprint feature is reported.

\subsection*{Structural Feature}
\begin{figure}[h]
  \centering
  {\includegraphics[width=6.0cm]{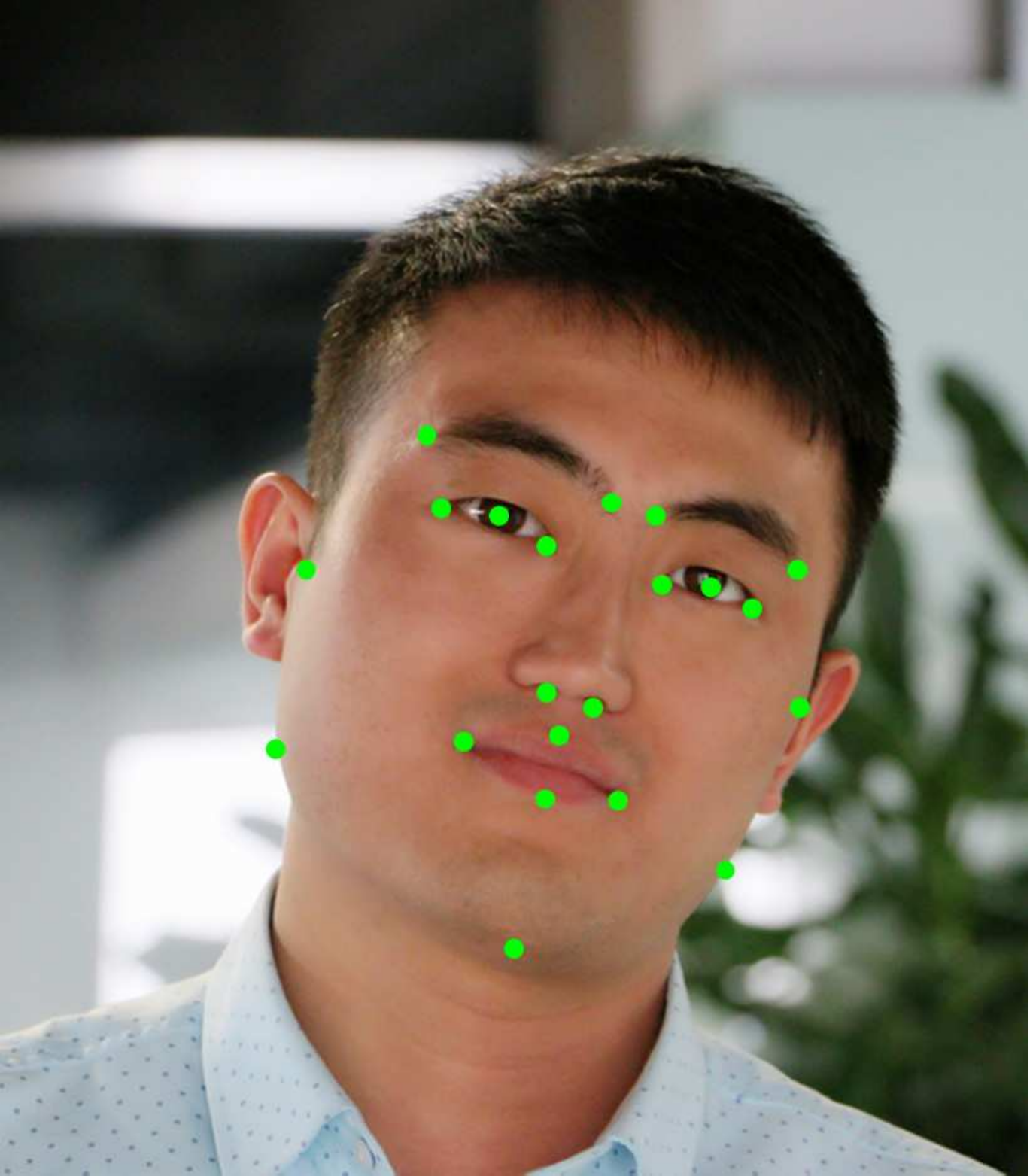}}
\caption{An example facial image and the 21 salient facial points detected by the LBF method}
\label{fig:2}
\end{figure}

To construct the structural feature of the face image, the salient facial points must first be detected. Many good facial point detectors have been described in the literature~\cite{bib9,bib23,bib24,bib25}; we used the LBF proposed in~\cite{bib9} in this work. Fig.~\ref{fig:2} shows an example of the detection result of this method. In~\cite{bib7}, Rojas et al. proposed to use the spatial relations of 21 specific salient facial points to construct a facial structural feature. In our work, to counter the possible effects of distance variations from the participants to the camera and variations in potential head pose, we normalized the coordinates of the 21 detected facial points by using a similarity transformation such that the two transformed pupils are horizontally positioned at a fixed distance.

The 21 detected points from each face image are denoted as $P = p_1, \dots, p_{21}\in R^2$, and their mean coordinates $M = m_1, \dots , m_{21}\in R^2$ are computed for the dataset. Finally, a 1134-dimensional structural feature vector of the face is constructed as follows:

\begin{enumerate}
\item The first 42 elements of the vector are the differences of the 21 points $p_i$ to their corresponding mean $m_i$ ($i = 1, \dots, 21$) in polar coordinates as also done in~\cite{bib7}.
\item The second set is an 882-D subvector encoding the spatial relations between each salient point $p_i$ of the face and all the points of the mean face image $m_i$ ($i, j = 1, \dots, 21$) in terms of radius and angle.
\item The third set is the 210 different point-to-point Euclidean distances encoding the intra face structural relationship.
\end{enumerate}

\subsection*{Appearance Feature}
\begin{figure}[h]
\begin{minipage}[b]{.23\linewidth}
  \centering
  \centerline{\includegraphics[width=2.5cm,height=2.8cm]{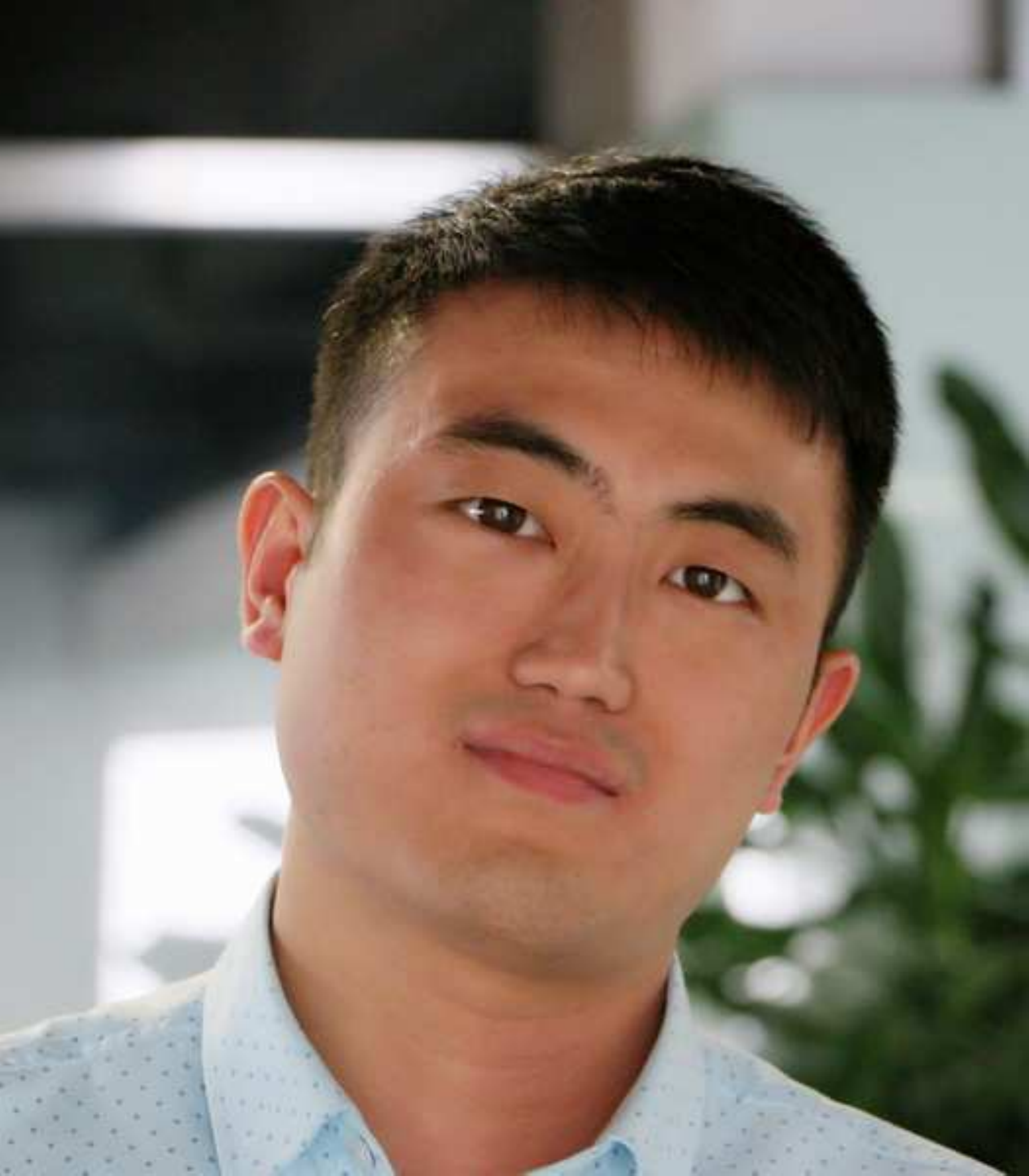}}
  \centerline{(a)}\medskip
\end{minipage}
\hfill
\begin{minipage}[b]{.23\linewidth}
  \centering
  \centerline{\includegraphics[width=2.5cm,height=2.8cm]{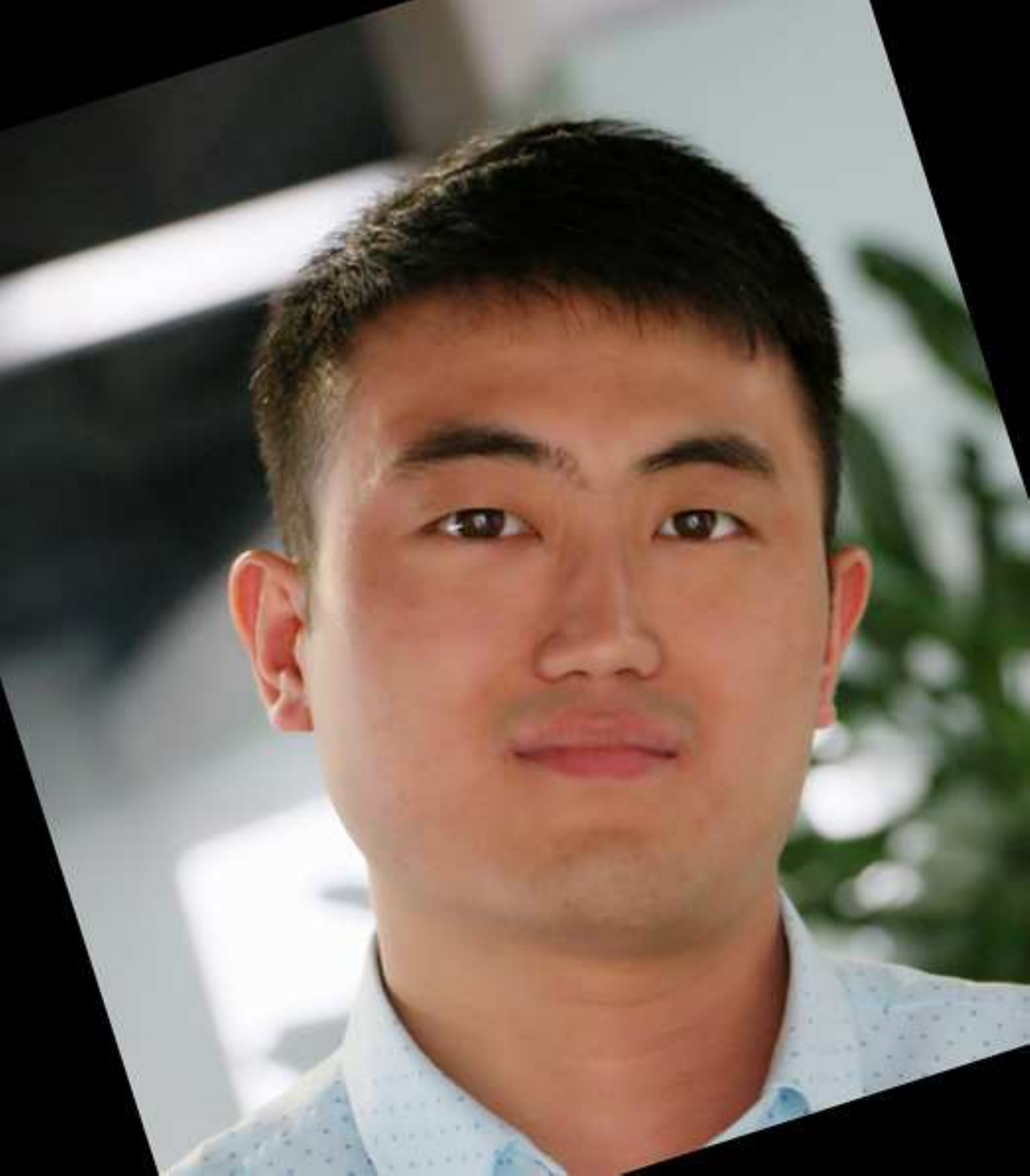}}
  \centerline{(b)}\medskip
\end{minipage}
\hfill
\begin{minipage}[b]{0.23\linewidth}
  \centering
  \centerline{\includegraphics[width=2.5cm,height=2.8cm]{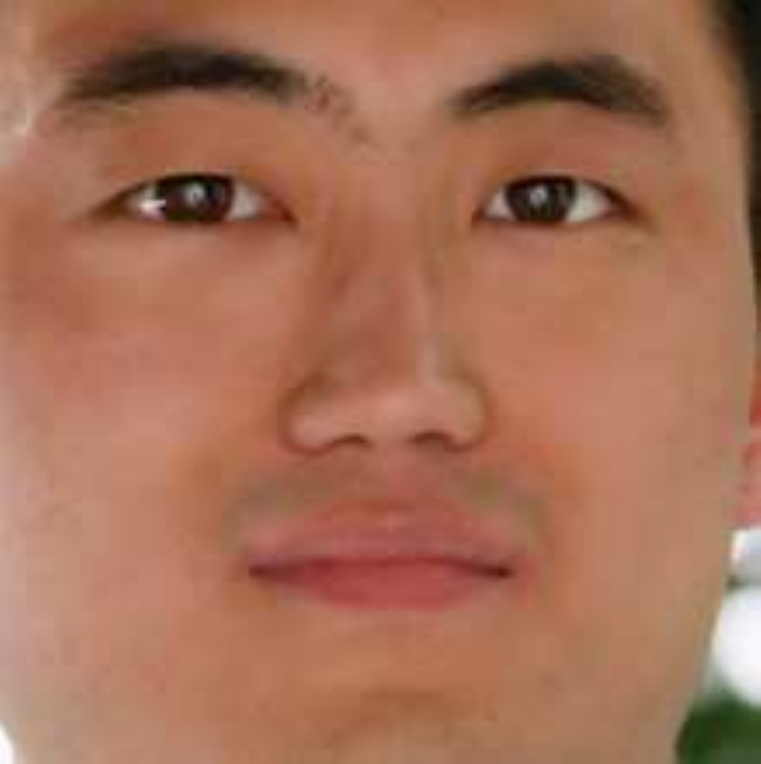}}
  \centerline{(c)}\medskip
\end{minipage}
\hfill
\begin{minipage}[b]{0.23\linewidth}
  \centering
  \centerline{\includegraphics[width=2.5cm,height=2.8cm]{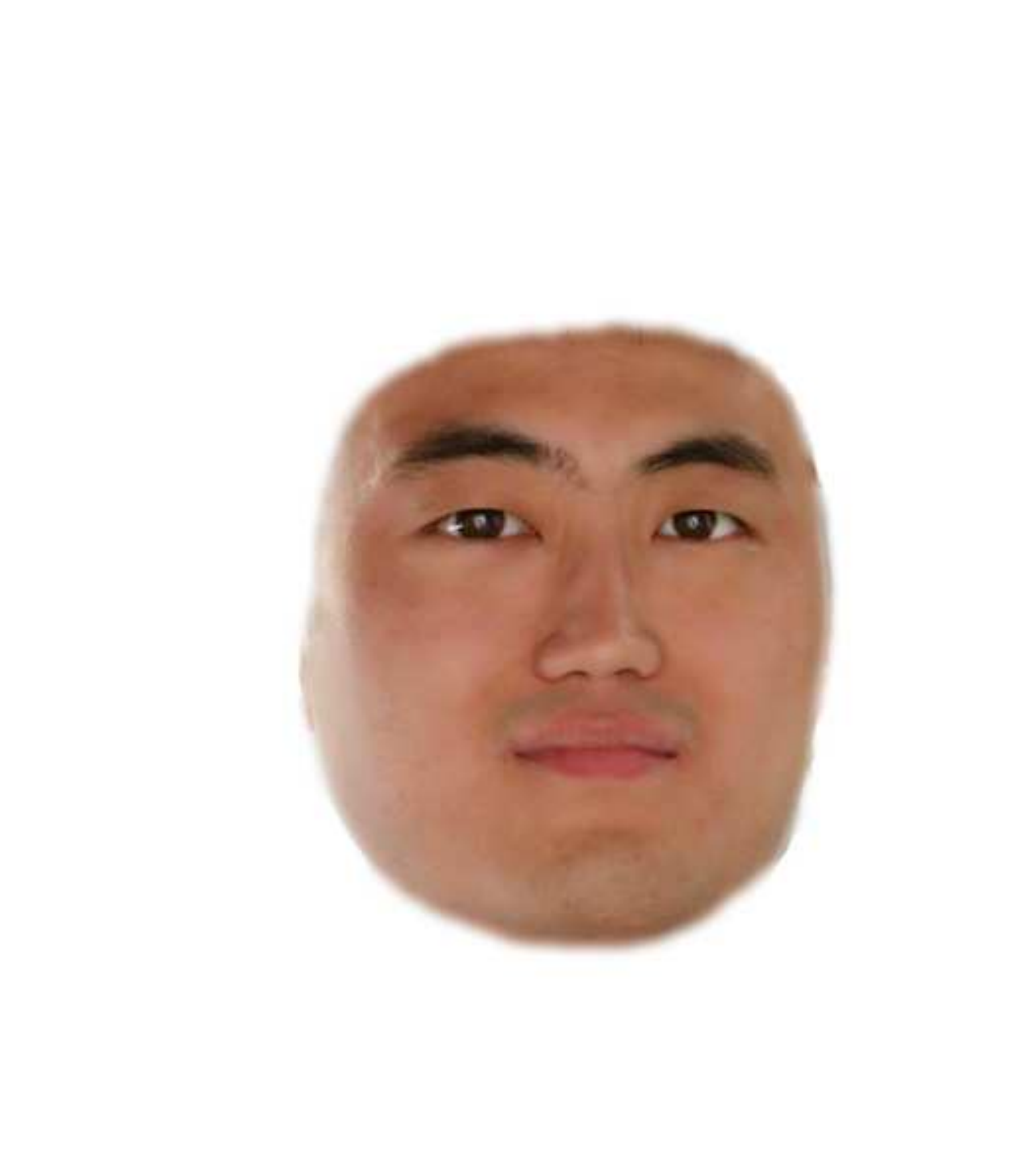}}
  \centerline{(d)}\medskip
\end{minipage}
\caption{Results at different preprocessing stages: (a) the original image, (b) the image after the similarity transformation, (c) the cropped image, and (d) the manually segmented image}
\label{fig:3}
\end{figure}

To study whether the structural information or the appearance information of the face is more informative for evaluating the self-reported personality traits and measured intelligence, we also extracted an appearance feature from the face image. More specifically, we first use the LBF~\cite{bib9} to detect the two pupil locations from each face image. Then, based on the coordinates of these two points, a similarity transformation (a rotation + rescaling in our work) is performed for all the images in the dataset such that the two transformed pupils are horizontally positioned with a fixed distance. Fig.~\ref{fig:3} (a) and (b) are an original image and its transformed image, respectively. In addition, because the photographed images have too much redundant background information, we select the image region to make sure the eyes lie horizontally at the same height and leave a standard length of neck visible. Prior to extracting the appearance feature of the facial images, the structural variations need to be standardized. In this work, the locations of the 21 salient points for each face were warped to their corresponding mean values to remove the shape variations.

\begin{figure}[h]
  \centering
  {\includegraphics[width=4.0cm]{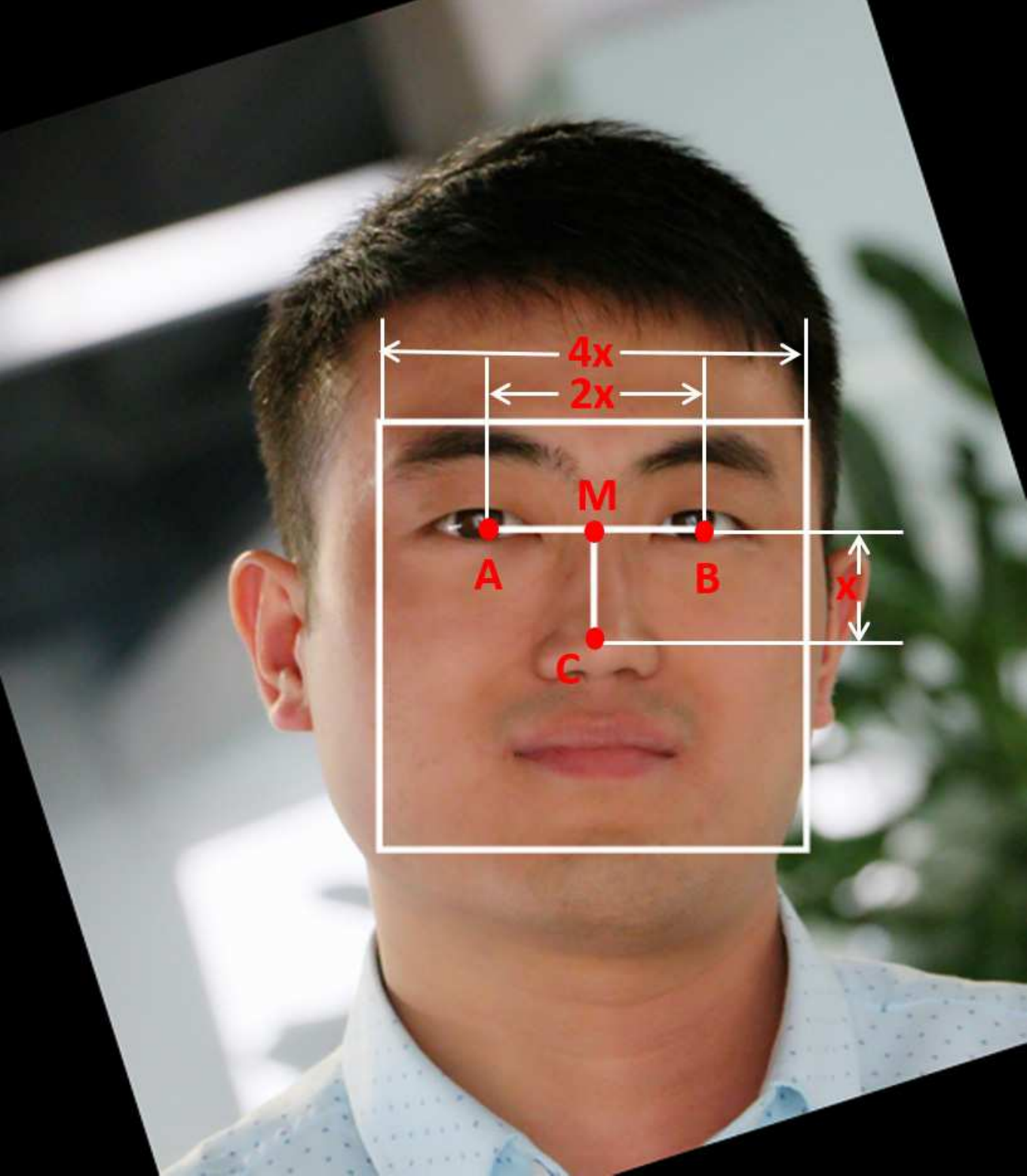}}
\caption{An illustration of the cropping proceduce for an image}
\label{fig:4}
\end{figure}

In facial analysis, researchers usually crop the image to remove irrelevant information such as the background, hair and clothing. In this work, the sample images are also cropped using the following steps (see Fig.~\ref{fig:4}). First, the two pupils are connected by a line segment AB, then a downwardly perpendicular segment MC is drawn where MC=1/2 AB and M is the midpoint of AB. Finally, the cropped region is a square with each side = 2AB and centered at C.  Fig.~\ref{fig:3} (c) shows an example of a cropped image. In addition, to further reduce the background influence, we manually segmented the facial images based on the contour of the face with Adobe Photoshop. Fig.~\ref{fig:3} (d) shows an example of a manually segmented image.

\begin{figure}[h]
  \centering
  {\includegraphics[width=12.5cm,height=4.0cm]{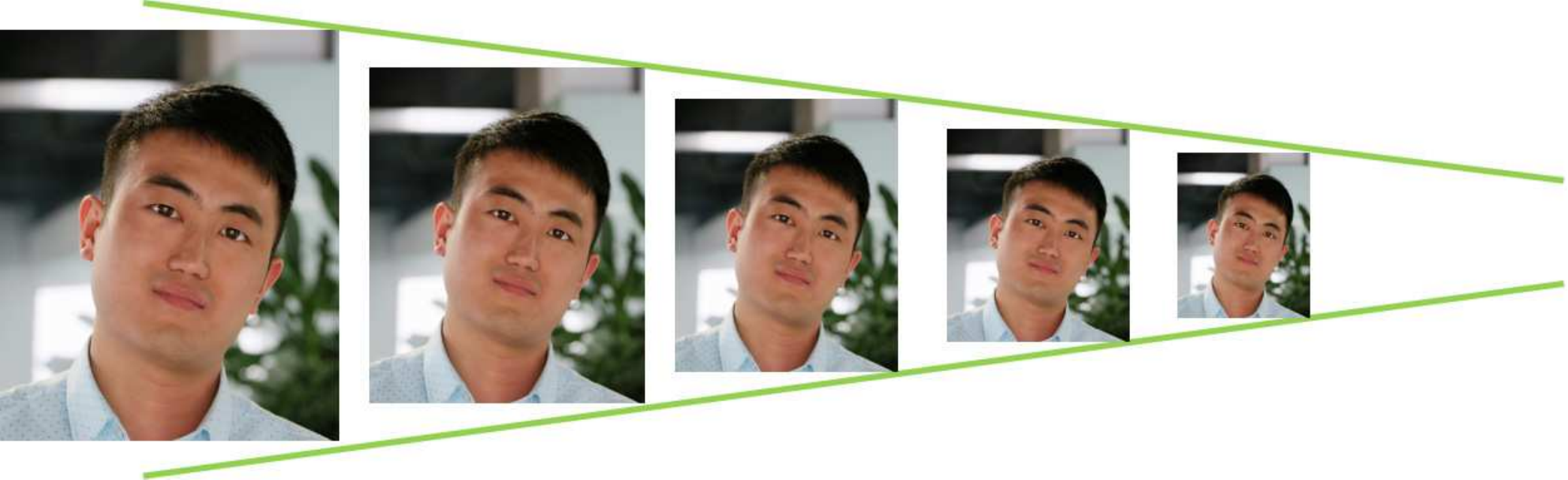}}
\caption{An example face image pyramid}
\label{fig:5}
\end{figure}

In the study of facial recognition, a face image pyramid is usually built to extract suitable appearance features such as LBP~\cite{bib26}, SIFT~\cite{bib27,bib28}, and Gist~\cite{bib29,bib30}. In this work, we extract not only global information but also local details from the face images at different scales. In addition, because we had few samples and dimensionality reduction of high dimensional representations usually can achieve better performance than a direct low-dimensional representation~\cite{bib31}, we also built an image pyramid for each facial image before extracting its appearance feature in this work. The scale parameter we used in our experiments is 1.5. Fig.~\ref{fig:5} shows an example of the face image pyramid.

In previous studies of facial recognition and face verification~\cite{bib31}, the features that are both discriminative for inter-person differentiation and invariant to intra-person variations are always employed. Recently, some handcrafted features have been designed and achieved good performance. In this work, we use five different descriptors to represent facial information. Four are local descriptors (HOG~\cite{bib10}, LBP~\cite{bib11}, Gabor~\cite{bib12}, and SIFT~\cite{bib14}), and one is a global descriptor (Gist~\cite{bib13}). For the Gabor and Gist descriptors, we used the implementation from the authors' home page; for the HOG, LBP and SIFT, we used an implementation of the open source library VLFeat~\cite{bib32}. When extracting the SIFT feature, we sorted all SIFT descriptors according to their scale values such that different images use the same number of descriptors to construct a fixed-length feature vector. Our final appearance feature is a concatenation of the results of the above five descriptors.

\subsection*{Fingerprint Feature}
\begin{figure}[h]
  \centering
  {\includegraphics[width=3.5cm,height=3.0cm]{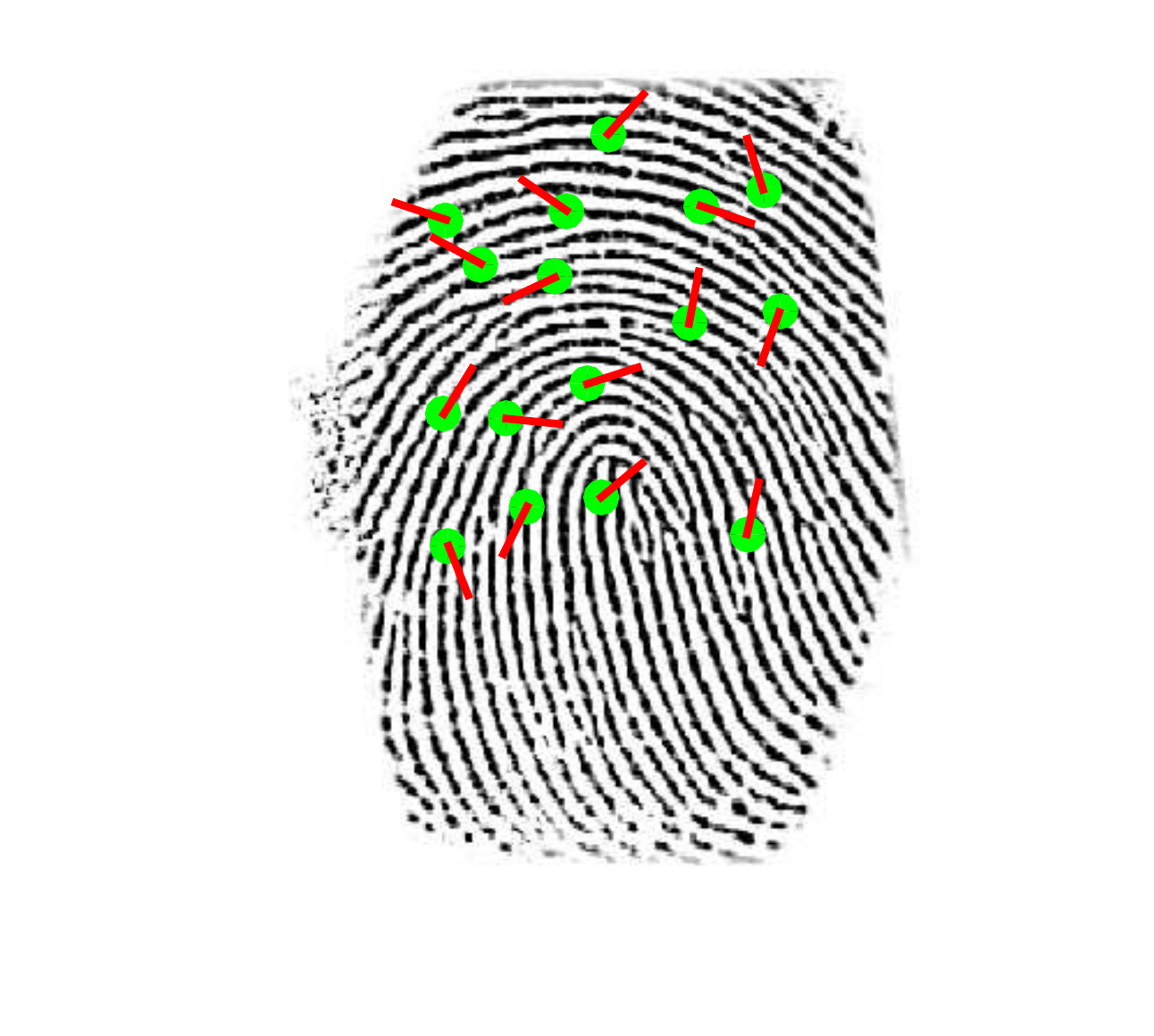}}
  {\includegraphics[width=3.5cm,height=3.0cm]{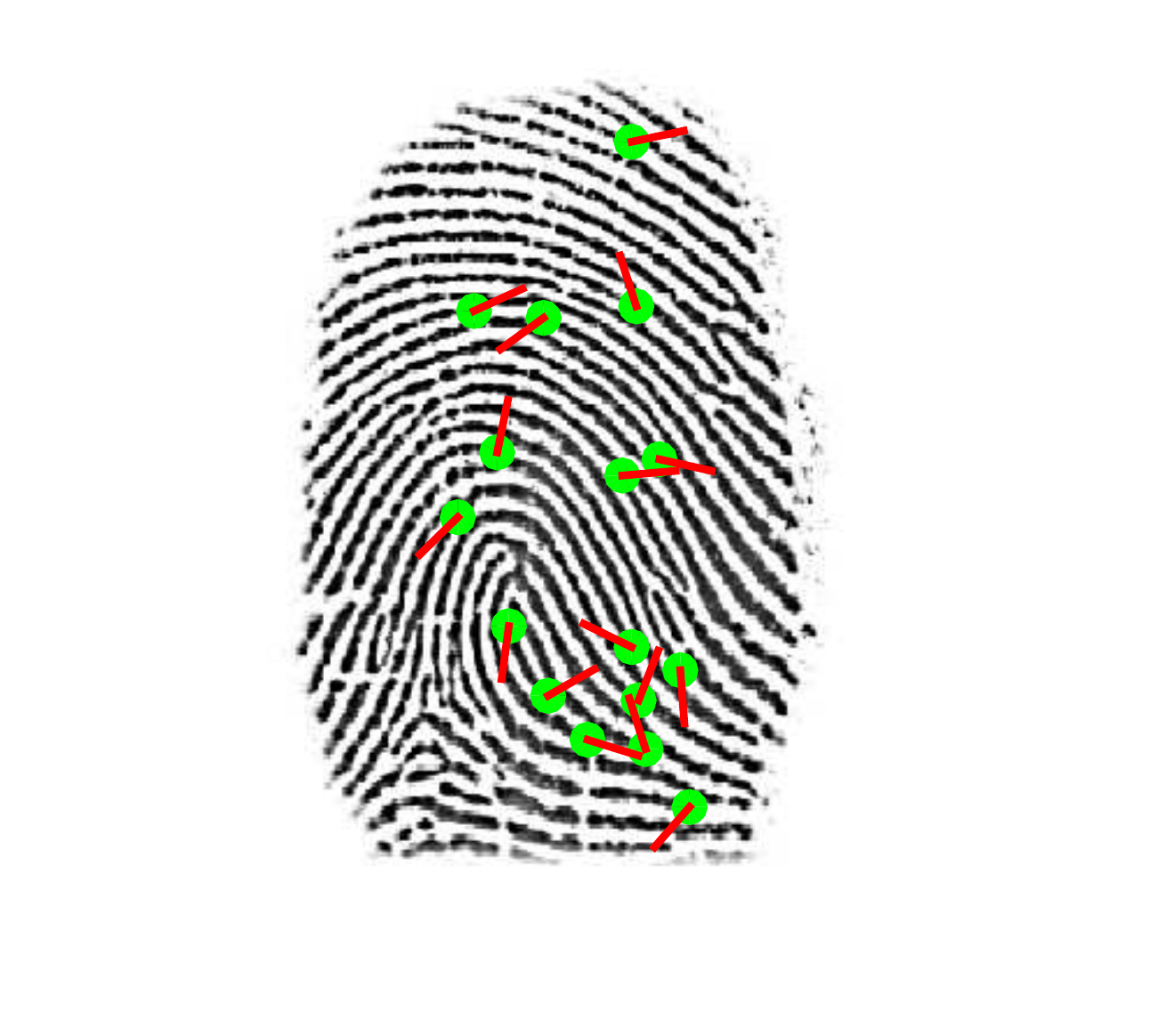}}
  {\includegraphics[width=3.5cm,height=3.0cm]{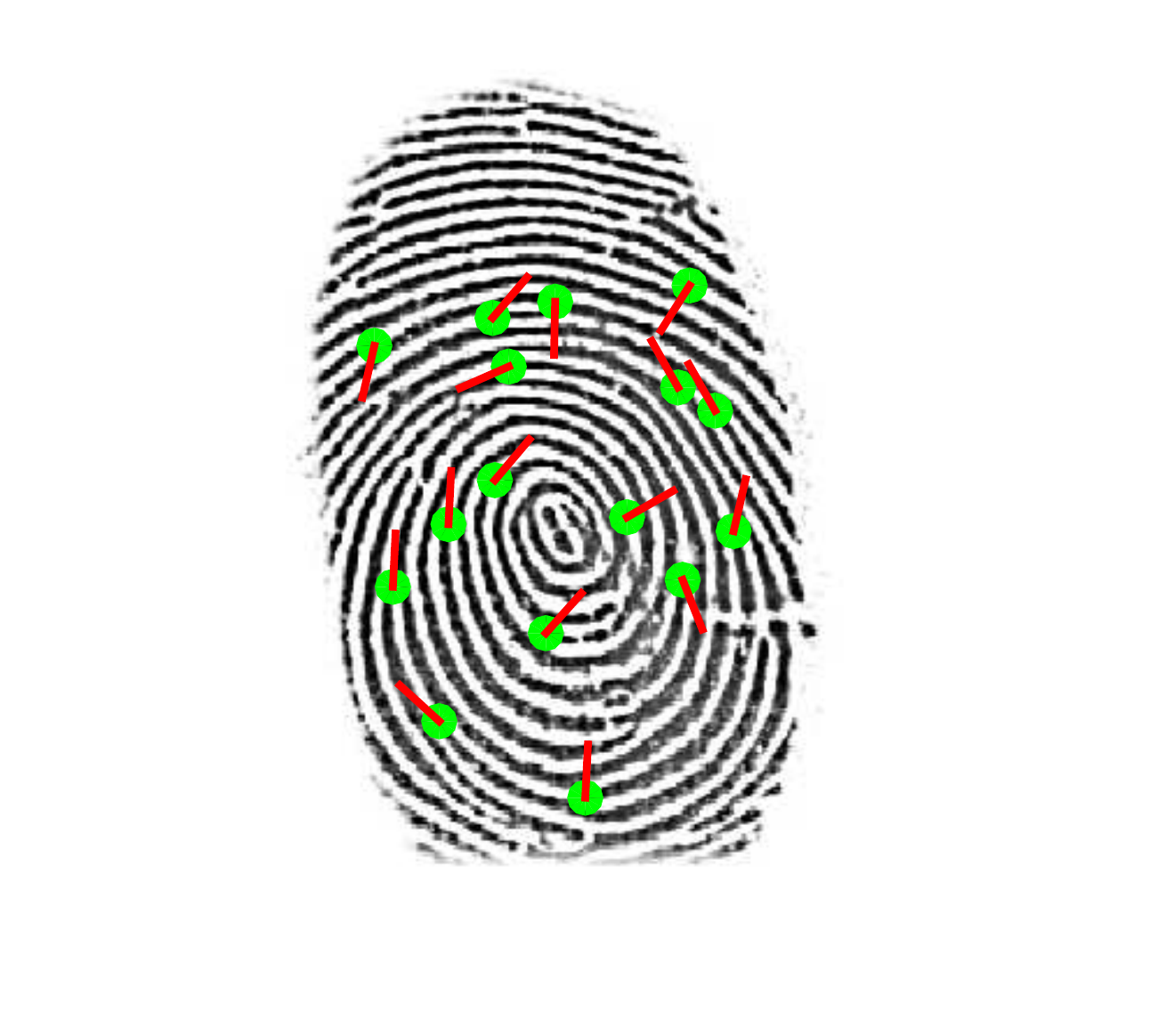}}
\caption{The fingerprint images and their detected minutiae and orientation}
\label{fig:6}
\end{figure}

Minutia-based fingerprint representation is widely used in fingerprint image matching. In this work, we also used a minutia-based representation for our fingerprint images. For each image, we selected 16 minutiae (determined by the number of minutiae) and used their locations and orientations to construct the fingerprint feature vector. Fig.~\ref{fig:6} shows three sample fingerprint images with detected minutiae.

\section*{Experiments and Results}
In this section, we performed our experiments using the constructed dataset and discuss the results. As described in the preceding sections, we obtained the following information:

\begin{enumerate}
\item The structural feature of the facial image for each sample
\item The appearance feature of the face image for each sample
\item The minutia-based feature of the fingerprint image for each sample
\item The discrete scores of 20 self-reported personality traits for each sample
\item The measured intelligence score for each sample.
\end{enumerate}

To investigate whether the self-reported personality traits and measured intelligence can be evaluated from the facial features and the fingerprint feature, we conducted classification and regression experiments, respectively. Considering that there are large differences in facial composition between men and women, we performed the experiments for men and women separately.

\subsection*{The Classification Experiments}
In our classification experiments, because the score on each personality trait measured by the 16PF is a discrete figure ranging from 1 to 10, we first binarized the personality traits by setting the highest 5 as the ``have trait" class and the lowest 5 as the ``do not have trait" class. Similarly, for the measured intelligence, we set the percentile scores less than or equal to 75$\%$ to one category and the scores greater than 75$\%$ to another category to balance the number of the samples in the two categories.

In this work, we use a bank of classifiers to study whether the self-reported personality traits and the measured intelligence can be predicted from the facial features accurately. Five widely used discriminative classification methods are selected to perform the experiments. They are Parzen Window~\cite{bib33}, Decision Tree~\cite{bib34}, K-Nearest Neighbor (KNN)~\cite{bib34}, Naive Bayes~\cite{bib35} and Random Forest~\cite{bib36}. For KNN, k is set to 5 following the suggestion in~\cite{bib8}. The implementations used for these classification methods are off-the-shelf routines from PRTools~\cite{bib37}.

Because our samples are not large, for each classifier used in our work, the accuracy is estimated with an N-fold cross-validation scheme and the training is repeated thirty times to obtain reliable standard deviations. We randomly divided the dataset into 10 non-overlapped subsets. Nine subsets were used for training while the remaining subset was used for testing. In our work, we used several criteria to evaluate the performance of each classification method. In addition to the classification accuracy, the confidence interval for a 95$\%$ confidence level is also computed to attest the reliability of these results, as follows:
\begin{equation}
I = 1.96 \cdot\sigma/\sqrt{N}
\label{equ:1}
\end{equation}
where $\sigma$ is the standard deviation of the results, and N is the number of repetitions using the cross validation framework.

As described above, we extracted two types of facial features from the face image. In the subsequent subsections, these features are used to train the classification models to predict the class labels of the personality traits and the measured intelligence, respectively.

\paragraph{A Discussion of the Structural Feature} Our structural feature has 1134 dimensions which is far more than the number of acquired samples; therefore, PCA was used to reduce the feature dimensionality. For any dimensionality reduction methods, choosing a suitable way to reduce the dimensions is a crucial issue. In our experiments, the dimensions of the original features are reduced to 5, 10, 15, 20, 25, 30, 35, 40 and 50 gradually, then the features of these different dimensions are used to train the five classifiers, respectively, to empirically determine the most appropriate feature dimensions.

\begin{figure}[h]
  \centering
  {\includegraphics[width=6.6cm,height=6.0cm]{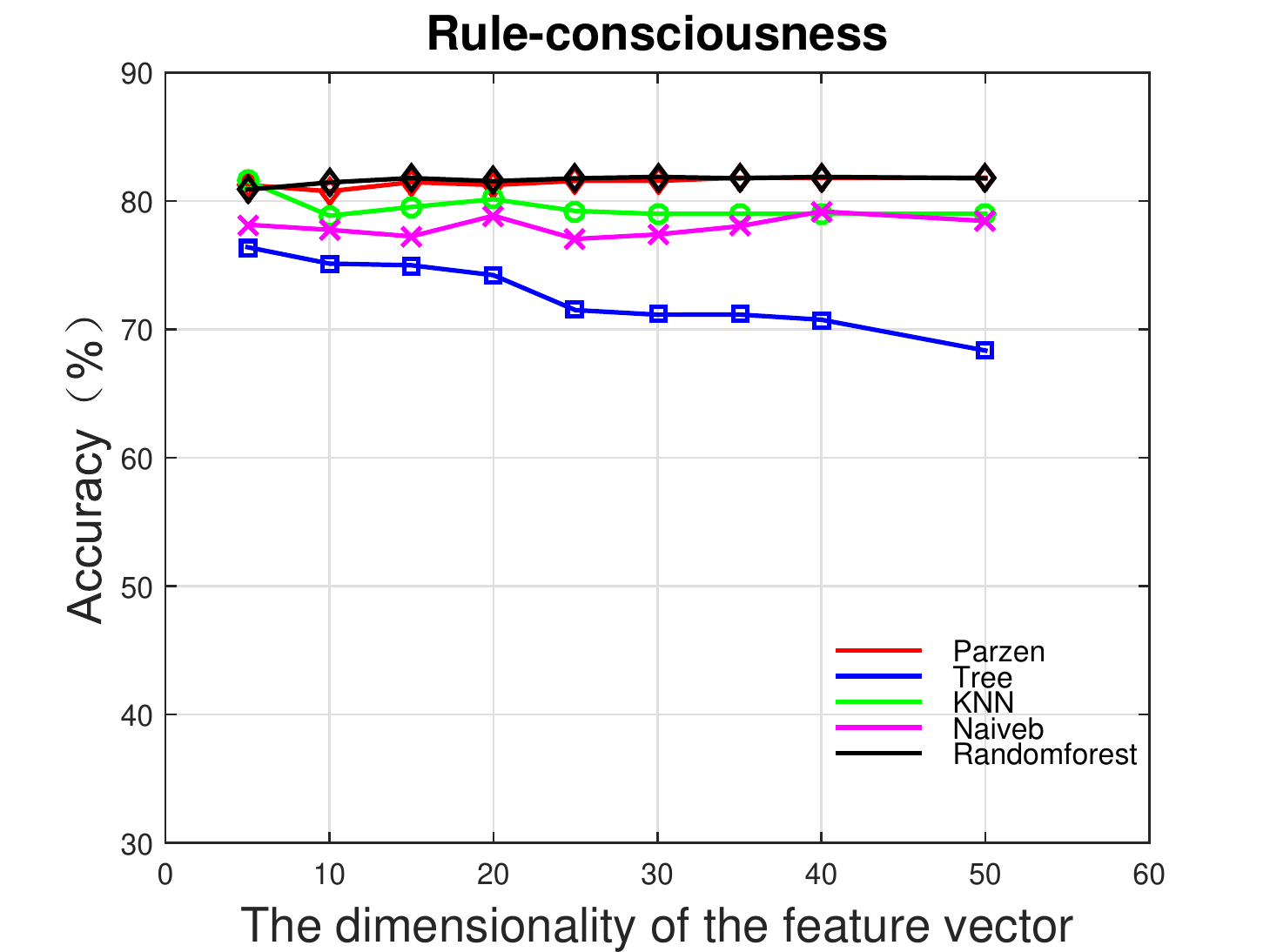}}
  {\includegraphics[width=6.6cm,height=6.0cm]{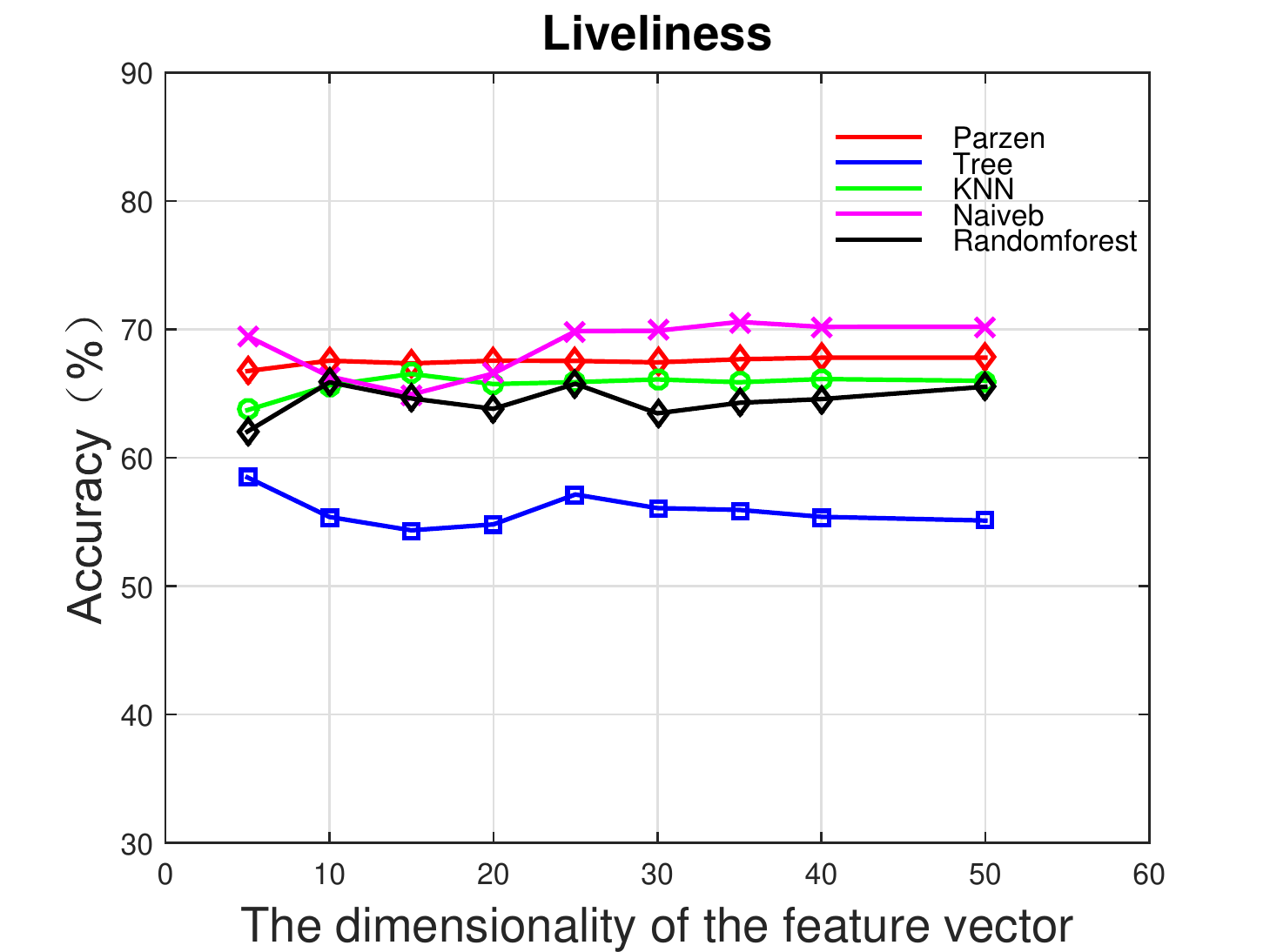}}
\caption{The classification accuracy of ``Rule-consciousness'' and ``Liveliness'' under different feature dimensions}
\label{fig:7}
\end{figure}

As shown in Fig.~\ref{fig:7}, the classification accuracy on the personality traits varies with respect to the reduced feature dimension. However, a clear variation trend does not exist for a specific classification method across all the personality traits. To select the dimension with the best performance for all the personality traits for a given classification method, we designed a dimension selection strategy as shown in Algorithm~\ref{alg:1}. For measured intelligence, because the problem is rather simple, we simply choose the dimension with the highest accuracy for a given classification method. Note that we also conduct experiments for both genders separately.

Our feature dimension selection procedure for a given classification method is shown in Algorithm~\ref{alg:1}:
\begin{algorithm}
\caption{Feature dimension selection procedure}
\begin{algorithmic}[1]
\Require For a given classification method:
\item Compute the classification accuracy for all the personality traits under different dimensions, denoted as:
    \begin{equation}
    m_i^j \quad (i=1, \dots,20; j=5,10,15,20,25,30,35,40,50)
    \label{equ:2}
    \end{equation}
    where $i$ is the index of personality traits, and $j$ is the reduced dimension index of the feature vector.
\item Find the maximal value of \{$m_i^j$\} for each personality trait as follows:
    \begin{equation}
    M_i = \max_{j = 5, 10, \dots, 50}m_i^j(i = 1, \dots, 20)
    \label{equ:3}
    \end{equation}
\item Normalize $m_i^j$ by $M_i$:
    \begin{equation}
    \bar{m}_i^j = \frac{m_i^j}{M_i}
    \label{equ:4}
    \end{equation}
\item Sum the normalized accuracy $\bar{m}_i^j$ of all the 20 personality traits with respect to a specific dimension:
    \begin{equation}
    N_j = \sum_{i=1}^{20}\bar{m}_j^i
    \label{equ:5}
    \end{equation}
\item Find the maximum among \{$N_j$\} whose corresponding dimension is the optimal one.
\end{algorithmic}
\label{alg:1}
\end{algorithm}

With the structural feature under the determined optimal dimension, we measured the classification accuracy of each classification method. Table~\ref{table1} and Table~\ref{table2} show the performance for all the five classification rules with respect to all the personality traits and intelligence (the optimal feature dimension of the five methods is shown in Table~\ref{table4}). To attest to the reliability of these results, the confidence interval is computed and shown in parenthesis. Note that the accuracy values for ``Rule-consciousness'' and ``Vigilance'' are well beyond chance levels for all the classification methods for both genders. In contrast, except for a few personality traits such as ``Stability'', ``Liveliness'' and ``Introversion/Extroversion'', other traits show near-chance predictability levels. In fact, the accuracy of some classification methods for these traits is below 50$\%$. The above results suggest that the personality traits ``Rule-consciousness'' and ``Vigilance'' are more related to facial morphological characteristics and that this relevance can also be recognized more accurately.
\begin{table}[!ht]
\footnotesize
\begin{adjustwidth}{-2.25in}{0in}
\caption{
{\bf Mean accuracy and the confidence interval (in parentheses) of the five classification rules for the 20 traits with respect to males and females. }}
\begin{tabular}{|l|l|l|l|l|l|l|l|l|l|l|}
\hline
\multicolumn{11}{|c|}{\bf Male-Structural feature}\\ \hline
\multicolumn{1}{|c|}{\bf Trait} & \multicolumn{1}{|c|}{\bf Warm} & \multicolumn{1}{|c|}{\bf Reas} & \multicolumn{1}{|c|}{\bf Stab} & \multicolumn{1}{|c|}{\bf Domin} & \multicolumn{1}{|c|}{\bf Live} & \multicolumn{1}{|c|}{\bf Cons} & \multicolumn{1}{|c|}{\bf Soci} & \multicolumn{1}{|c|}{\bf Sens} & \multicolumn{1}{|c|}{\bf Vigil} & \multicolumn{1}{|c|}{\bf Abst}\\ \hline
\bf Parzen & 52.87(2.6) & 55.18(1.5) & 65.79(2.2) & 55.61(1.4) & 67.43(2.1) & 81.56(1.7) & 56.61(1.2) & 39.18(1.8) & 72.64(2.3) & 63.36(1.1)\\ \hline
\bf DTree & 48.60(3.5) & 53.68(4.5) & 60.69(2.6) & 54.01(2.3) & 56.08(2.2) & 71.14(2.4) & 44.78(3.0) & 53.33(2.7) & 59.17(2.3) & 49.90(2.6)\\ \hline
\bf KNN & 58.40(1.5) & 49.94(3.2) & 60.49(2.0) & 50.83(2.3) & 65.74(1.1) & 80.13(1.6) & 53.75(1.8) & 38.63(2.6) & 69.57(0.7) & 62.04(0.7)\\ \hline
\bf NaiveB & 43.35(2.5) & 52.78(2.3) & 64.75(1.8) & 54.88(4.0) & 69.85(1.5) & 77.04(1.1) & 46.28(1.0) & 42.36(2.3) & 68.57(1.5) & 52.89(1.9)\\ \hline
\bf RF & 52.50(2.4) & 54.10(2.1) & 64.79(2.3) &	54.31(3.0) & 63.81(1.7) & 81.54(1.8) & 48.28(2.5) & 47.06(2.1) & 72.00(1.9)  & 57.75(2.1)\\ \hline
\multicolumn{11}{|c|}{}\\ \hline
\multicolumn{1}{|c|}{\bf Trait} & \multicolumn{1}{|c|}{\bf Priv} & \multicolumn{1}{|c|}{\bf Appr} & \multicolumn{1}{|c|}{\bf Open} & \multicolumn{1}{|c|}{\bf Reli} & \multicolumn{1}{|c|}{\bf Perf} & \multicolumn{1}{|c|}{\bf Tens} & \multicolumn{1}{|c|}{\bf Adap} & \multicolumn{1}{|c|}{\bf Intro} & \multicolumn{1}{|c|}{\bf Impet} & \multicolumn{1}{|c|}{\bf Cowa}\\ \hline
\bf Parzen & 66.93(1.4) & 65.47(1.2) & 39.60(3.1) & 53.36(1.7) & 56.90(1.6)  & 63.65(1.3) & 59.53(2.0) & 67.36(1.1) & 55.21(1.3) & 52.90(1.3)\\ \hline
\bf DTree & 56.68(2.3) & 51.99(2.6) & 49.64(2.2) & 50.82(2.0) & 56.32(2.2) & 52.21(2.4) & 56.31(2.1) & 55.61(1.1) & 55.13(2.2) & 51.58(2.3)\\ \hline
\bf KNN & 66.50(1.8)  & 61.42(1.3) & 46.44(2.0) & 52.14(2.3)& 51.47(2.6) & 58.93(1.9) & 53.88(2.1) & 60.69(1.2) & 57.92(1.7) & 48.60(2.7)\\ \hline
\bf NaiveB & 54.94(2.4) & 62.62(2.5) & 50.74(2.8) & 48.76(1.7) & 53.76(2.6) & 49.92(2.3) & 52.50(1.1)  & 59.17(1.7) & 49.89(2.1) & 47.36(2.3)\\ \hline
\bf RF & 64.18(1.6) & 65.78(2.1) & 56.68(2.5) & 52.07(2.1) & 53.17(2.4) & 57.47(2.1) & 51.71(1.8) & 62.47(1.5) & 51.99(2.5) & 51.89(2.1)\\ \hline
\multicolumn{11}{|c|}{\bf Female-Structural feature}\\ \hline
\multicolumn{1}{|c|}{\bf Trait} & \multicolumn{1}{|c|}{\bf Warm} & \multicolumn{1}{|c|}{\bf Reas} & \multicolumn{1}{|c|}{\bf Stab} & \multicolumn{1}{|c|}{\bf Domin} & \multicolumn{1}{|c|}{\bf Live} & \multicolumn{1}{|c|}{\bf Cons} & \multicolumn{1}{|c|}{\bf Soci} & \multicolumn{1}{|c|}{\bf Sens} & \multicolumn{1}{|c|}{\bf Vigil} & \multicolumn{1}{|c|}{\bf Abst}\\ \hline
\bf Parzen & 54.22(1.3) & 51.78(2.6) & 70.00(2.3) & 70.67(1.1) & 76.67(1.3) & 82.22(2.1) & 44.11(1.6) & 56.89(1.1) & 82.22(1.5) & 69.22(1.3)\\ \hline
\bf DTree & 49.67(2.1) & 53.89(2.9) & 59.33(3.3) & 56.67(2.9) & 65.33(1.5) & 67.44(3.0) & 45.56(2.6) & 52.11(2.9) & 69.78(2.2) & 55.89(2.8)\\ \hline
\bf KNN & 49.78(3.0) & 48.78(4.0) & 68.00(2.7) & 67.78(1.7) & 75.33(1.3) & 81.44(1.3) & 61.11(1.8) & 53.67(2.3) & 80.78(1.3) & 67.44(1.4)\\ \hline
\bf NaiveB & 55.33(2.7) & 60.89(1.9) & 62.22(2.0) & 73.67(1.6) & 69.56(1.7) & 77.22(1.7) & 48.33(2.1) & 49.22(1.5) & 69.89(1.2) & 64.00(1.5)\\ \hline
\bf RF & 55.89(1.7) & 54.33(2.1) & 67.89(1.6) & 70.56(2.5) & 74.11(1.4) & 82.22(2.3) & 49.67(2.3) & 57.22(2.1) & 81.56(1.6) & 65.89(2.1)\\ \hline
\multicolumn{11}{|c|}{}\\ \hline
\multicolumn{1}{|c|}{\bf Trait} & \multicolumn{1}{|c|}{\bf Priv} & \multicolumn{1}{|c|}{\bf Appr} & \multicolumn{1}{|c|}{\bf Open} & \multicolumn{1}{|c|}{\bf Reli} & \multicolumn{1}{|c|}{\bf Perf} & \multicolumn{1}{|c|}{\bf Tens} & \multicolumn{1}{|c|}{\bf Adap} & \multicolumn{1}{|c|}{\bf Intro} & \multicolumn{1}{|c|}{\bf Impet} & \multicolumn{1}{|c|}{\bf Cowa}\\ \hline
\bf Parzen & 54.11(1.4) & 53.78(2.4) & 60.11(1.7) & 60.56(1.2) & 58.11(1.0) & 54.44(1.3) & 66.56(1.5) & 65.44(1.3) & 75.56(2.2) & 71.11(1.8)\\ \hline
\bf DTree & 46.11(2.4) & 52.11(1.7) & 50.22(2.8) & 46.22(1.8) & 53.00(2.9) & 56.11(2.2) & 58.11(2.1) & 51.33(1.8) & 63.44(3.6) & 60.11(3.0)\\ \hline
\bf KNN & 50.00(2.0) & 53.33(1.5) & 57.89(2.0) & 57.11(1.6) & 54.56(2.3) & 49.11(2.3) & 64.78(2.7) & 63.89(1.2) & 73.67(2.6) & 70.00(2.7)\\ \hline
\bf NaiveB & 47.67(2.4) & 58.22(1.6) & 59.22(1.7) & 52.78(1.7) & 54.11(3.0) & 52.44(2.7) & 62.56(2.1) & 57.78(1.5) & 66.22(1.5) & 66.33(2.4)\\ \hline
\bf RF & 50.67(2.1) & 57.22(2.3) & 55.22(1.9) & 58.00(1.5) & 60.22(2.3) & 54.11(2.1) & 56.11(2.4) & 64.89(1.7) & 49.44(2.3) & 47.11(2.5)\\ \hline
\end{tabular}
\label{table1}
\end{adjustwidth}
\end{table}

\begin{table}[!ht]
\footnotesize
\caption{
{\bf Mean accuracy and the confidence interval (in parentheses) of the five classification rules for intelligence based on the structural and appearance features with respect to males and the females.}}
\begin{tabular}{|l|l|l|l|l|l|l|}
\hline
\multicolumn{1}{|c|}{\bf Feature} & \multicolumn{2}{|c|}{\bf Structural} & \multicolumn{4}{|c|}{\bf Appearance}\\ \hline
\multicolumn{1}{|c|}{\bf Processing} & {} & {} & \multicolumn{2}{|c|}{\bf Cropping} & \multicolumn{2}{|c|}{\bf Segmentation}\\ \hline \multicolumn{1}{|c|}{\bf Gender} & \multicolumn{1}{|c|}{\bf Male} & \multicolumn{1}{|c|}{\bf Female} & \multicolumn{1}{|c|}{\bf Male} & \multicolumn{1}{|c|}{\bf Female} & \multicolumn{1}{|c|}{\bf Male} & \multicolumn{1}{|c|}{\bf Female}\\ \hline
\bf Parzen & 47.39(2.0) & 56.33(2.1) & 51.00(2.3) & 58.89(1.7) & 53.04(2.7) & 50.89(2.1)\\ \hline
\bf DTree & 62.39(2.4) & 56.78(2.5) & 51.28(3.1) & 58.56(2.1) & 49.53(2.3) & 50.22(2.0)\\ \hline
\bf KNN & 49.89(1.9) & 59.11(2.4) & 53.40(1.5) & 61.89(2.5) & 47.19(2.0) & 57.89(1.6)\\ \hline
\bf NaiveB & 49.01(2.5) & 54.78(3.3) & 42.13(1.9) & 64.11(1.3) & 50.04(1.9) & 50.11(2.8)\\ \hline
\bf RF & 51.10(2.3) & 54.11(2.7) & 45.13(2.5) & 57.56(2.2) & 49.42(2.5) & 51.44(2.3)\\ \hline
\end{tabular}
\label{table2}
\end{table}

Note that the prediction accuracies for most of the personality traits for women are clearly higher than for men. However, the accuracies for the traits ``Privateness'' and ``Apprehension'' for women are lower than those for men. Thus, these two traits may be more related to men's facial features. These results suggest that due to the large differences in facial composition between men and women, some personality traits can be predicted from men's facial images more accurately while other traits can be predicted from women's facial images more accurately.

As for the measured intelligence, the results in Table~\ref{table2} show that, for men, all five classification methods achieve only near-chance prediction levels while, for women, the methods achieve only slightly higher than chance levels. This suggests that the structural feature is not well suited for classifying measured intelligence.

There are some facts and research findings that can help interpret the above results. In the psychological field, researchers have performed numerous comparative experiments among large numbers of twins and found that approximately 50$\%$ of a human's personality traits are affected by genetics. Some personality traits depend largely on genetic qualities and can be little influenced by practice or life experiences, while some other personality traits mainly depend on the social environment. At the same time, biological studies demonstrate that humans' facial characteristics are to a large extent genetically determined. Therefore, the traits more related to genetic factors may have a stronger correlation with facial features (such as ''Rule-consciousness'' and ``Vigilance'' in this work) and can be predicted from the face images more accurately.

\paragraph{A Discussion of the Appearance Feature} In our work, five texture descriptors (HOG, LBP, Gabor, Gist and SIFT) were used to extract the facial appearance information. Considering that different descriptors may extract complementary information for predicting personality traits and intelligence, we used PCA to reduce the results by the five descriptors to a lower dimension first (85 dimensions in our work), and then, concatenated them into a single holistic appearance feature. Similar to the experiments based on the structural feature, the dimensions of this holistic appearance feature were reduced to 5, 10, 15, 20, 25, 30, 40, 50, 60, 70 and 80 gradually using PCA. Then, for all the personality traits, the optimal dimension for each classification method was selected according to Algorithm~\ref{alg:1}. With the appearance feature under the obtained optimal dimension, we measured classification accuracy of each classification method with respect to all the personality traits and intelligence. As described in the preceding section, image cropping and segmenting were first carried out for the facial images to remove irrelevant information. Here, we perform the experiments on the cropped and segmented facial images separately.

Table~\ref{table3} and Table~\ref{table2} show the results of all five classification methods with respect to each personality trait and intelligence. The confidence interval is shown in parenthesis. Note that the accuracy scores for ``Rule-consciousness and ``Vigilance'' are still well beyond chance levels for all the classification methods. For men, the prediction accuracy of ``Rule-consciousness'' for all the methods except the Decision Tree is higher than 74$\%$. The accuracies of ``Vigilance'' for the Parzen Window and KNN methods are higher than 71$\%$; however, the accuracies for the Naive Bayes and the Random Forest methods are lower than 70$\%$ but still higher than 61$\%$. For women, the results were better. For both ``Rule-consciousness'' and ``Vigilance'' for all classification methods except the Decision Tree, the accuracy is higher than 74$\%$. Moreover, the performances of the Naive Bayes and the Random Forest methods for ``Vigilance'' are much better than for men, and both are higher than 74$\%$. The consistent results on ``Rule-consciousness'' and ``Vigilance'' from both the structural feature and appearance feature suggest that these two personality traits may have strong correlations with the facial characteristics of both men and women.
\begin{table}[!ht]
\footnotesize
\begin{adjustwidth}{-2.25in}{0in}
\caption{
{\bf Mean accuracy and the confidence interval (in parentheses) of the five classification rules for the 20 traits with respect to both genders, for cropped and segmented images.}}
\begin{tabular}{|l|l|l|l|l|l|l|l|l|l|l|}
\hline
\multicolumn{11}{|c|}{\bf Male-Appearance feature-Cropped image}\\ \hline
\multicolumn{1}{|c|}{\bf Trait} & \multicolumn{1}{|c|}{\bf Warm} & \multicolumn{1}{|c|}{\bf Reas} & \multicolumn{1}{|c|}{\bf Stab} & \multicolumn{1}{|c|}{\bf Domin} & \multicolumn{1}{|c|}{\bf Live} & \multicolumn{1}{|c|}{\bf Cons} & \multicolumn{1}{|c|}{\bf Soci} & \multicolumn{1}{|c|}{\bf Sens} & \multicolumn{1}{|c|}{\bf Vigil} & \multicolumn{1}{|c|}{\bf Abst}\\ \hline
\bf Parzen & 50.44(1.5) & 52.07(1.0) & 64.32(1.3) & 53.65(2.2) & 66.26(1.1) & 81.18(1.0) & 55.65(1.0) & 46.31(2.0) & 72.50(1.0) & 61.61(1.5)\\ \hline
\bf DTree & 50.36(2.6) & 52.72(2.7) & 55.32(2.4) & 54.14(1.0) & 62.53(2.0) & 75.44(2.4) & 53.96(3.2) & 53.69(2.2) & 62.40(2.5) & 60.00(2.5)\\ \hline
\bf KNN & 54.57(1.9) & 56.71(2.7) & 59.67(2.4) & 47.78(1.0) & 67.33(2.0) & 81.18(2.4) & 49.18(3.2) & 43.82(2.2) & 71.83(2.5) & 61.96(2.5)\\ \hline
\bf NaiveB & 52.10(2.3) & 53.04(2.9) & 59.04(1.9) & 64.13(2.0) & 61.13(1.7) & 74.35(1.4) & 63.96(1.4) & 61.04(2.1) & 66.74(2.0) & 59.47(2.5)\\ \hline
\bf RF & 50.99(2.1) & 49.06(2.5) & 62.10(1.1) & 50.79(1.5) & 63.00(1.3) & 80.74(2.1) & 55.56(2.5) & 50.78(2.3) & 68.29(1.9) & 61.58(2.1)\\ \hline
\multicolumn{11}{|c|}{}\\ \hline
\multicolumn{1}{|c|}{\bf Trait} & \multicolumn{1}{|c|}{\bf Priv} & \multicolumn{1}{|c|}{\bf Appr} & \multicolumn{1}{|c|}{\bf Open} & \multicolumn{1}{|c|}{\bf Reli} & \multicolumn{1}{|c|}{\bf Perf} & \multicolumn{1}{|c|}{\bf Tens} & \multicolumn{1}{|c|}{\bf Adap} & \multicolumn{1}{|c|}{\bf Intro} & \multicolumn{1}{|c|}{\bf Impet} & \multicolumn{1}{|c|}{\bf Cowa}\\ \hline
\bf Parzen & 65.92(1.5) & 64.92(1.2) & 45.89(2.1) & 55.08(1.6) & 55.49(2.1) & 60.94(1.8) & 57.24(1.8) & 66.19(1.9) & 54.71(1.0) & 46.10(1.8)\\ \hline
\bf DTree & 53.76(2.7) & 54.60(2.6) & 49.97(4.3) & 60.74(3.0) & 52.07(2.9) & 56.72(1.7) & 50.36(3.2) & 54.53(3.0) & 46.92(2.6) & 50.93(3.3)\\ \hline
\bf KNN & 62.19(1.2) & 62.85(1.0) & 55.14(4.2) & 59.75(2.0) & 50.71(3.5) & 62.10(1.2) & 55.24(3.0) & 62.92(1.7) & 53.03(2.5) & 52.93(2.6)\\ \hline
\bf NaiveB & 54.94(1.8) & 58.15(1.4) & 45.56(1.5) & 60.68(2.1) & 58.22(2.6) & 52.72(2.8) & 55.94(2.3) & 64.31(2.5) & 54.72(2.2) & 44.01(2.0)\\ \hline
\bf RF & 63.04(2.3) & 63.43(1.8) & 49.76(2.7) & 58.81(2.5) & 52.51(2.5) & 56.17(2.1) & 55.83(2.7) & 63.18(1.1) & 52.78(2.9) & 47.68(2.3)\\ \hline
\multicolumn{11}{|c|}{\bf Male-Appearance feature-Segmented image}\\ \hline
\multicolumn{1}{|c|}{\bf Trait} & \multicolumn{1}{|c|}{\bf Warm} & \multicolumn{1}{|c|}{\bf Reas} & \multicolumn{1}{|c|}{\bf Stab} & \multicolumn{1}{|c|}{\bf Domin} & \multicolumn{1}{|c|}{\bf Live} & \multicolumn{1}{|c|}{\bf Cons} & \multicolumn{1}{|c|}{\bf Soci} & \multicolumn{1}{|c|}{\bf Sens} & \multicolumn{1}{|c|}{\bf Vigil} & \multicolumn{1}{|c|}{\bf Abst}\\ \hline
\bf Parzen & 50.06(3.3) & 48.25(1.6) & 65.64(2.0) & 52.28(3.1) & 68.82(2.4) & 81.68(1.5) & 59.60(2.8) & 44.44(1.8) & 72.63(2.2) & 61.68(1.9)\\ \hline
\bf DTree & 58.60(2.3) & 46.79(1.9) & 55.42(2.1) & 48.46(2.5) & 55.39(1.5) & 66.76(1.3) & 55.81(2.1) & 59.69(2.1) & 54.04(1.3) & 58.76(2.5)\\ \hline
\bf KNN & 52.43(2.6) & 55.11(1.6) & 64.53(1.9) & 51.63(1.7) & 68.24(2.2) & 81.57(1.8) & 53.96(1.7) & 58.36(2.0) & 72.17(1.7) & 61.79(1.4)\\ \hline
\bf NaiveB & 46.99(2.7) & 47.76(1.0) & 57.36(1.0) & 46.88(2.4) & 70.88(1.2) & 78.89(1.0) & 65.88(2.2) & 49.67(1.5) & 62.44(1.1) & 48.04(1.0)\\ \hline
\bf RF & 57.50(1.7) & 46.04(1.6) & 60.33(2.2) & 57.04(2.2) & 61.11(2.6) & 81.32(3.1) & 53.44(2.2) & 56.33(2.1) & 70.01(3.4) & 61.17(1.8)\\ \hline
\multicolumn{11}{|c|}{}\\ \hline
\multicolumn{1}{|c|}{\bf Trait} & \multicolumn{1}{|c|}{\bf Priv} & \multicolumn{1}{|c|}{\bf Appr} & \multicolumn{1}{|c|}{\bf Open} & \multicolumn{1}{|c|}{\bf Reli} & \multicolumn{1}{|c|}{\bf Perf} & \multicolumn{1}{|c|}{\bf Tens} & \multicolumn{1}{|c|}{\bf Adap} & \multicolumn{1}{|c|}{\bf Intro} & \multicolumn{1}{|c|}{\bf Impet} & \multicolumn{1}{|c|}{\bf Cowa}\\ \hline
\bf Parzen & 69.69(1.8) & 65.97(1.0) & 49.90(2.5) & 50.96(1.9) & 53.86(1.0) & 62.25(1.3) & 50.35(2.4) & 66.90(1.4) & 55.47(2.1) & 55.89(2.9)\\ \hline
\bf DTree & 56.43(1.4) & 55.51(2.5) & 51.93(2.3) & 49.78(1.9) & 50.92(2.5) & 52.53(1.9) & 56.21(2.5) & 55.88(1.5) & 55.50(1.3) & 48.60(2.1)\\ \hline
\bf KNN & 64.71(2.0) & 62.63(2.3) & 46.97(1.9) & 52.10(3.0) & 47.18(2.7) & 60.82(2.7) & 49.44(1.1) & 65.22(1.9) & 54.61(2.3) & 57.33(2.7)\\ \hline
\bf NaiveB & 61.44(1.3) & 61.33(2.0) & 47.85(3.0) & 51.49(2.5) & 49.96(2.1) & 54.15(1.4) & 52.24(1.9) & 73.00(1.1) & 57.72(2.0) & 56.44(1.6)\\ \hline
\bf RF & 65.18(3.6) & 65.58(1.9) & 48.07(3.3) & 47.71(2.9) & 50.79(1.3) & 59.90(2.1) & 54.06(2.6) & 61.33(3.7) & 57.07(2.3) & 50.08(2.0)\\ \hline
\multicolumn{11}{|c|}{\bf Female-Appearance feature-Cropped image}\\ \hline
\multicolumn{1}{|c|}{\bf Trait} & \multicolumn{1}{|c|}{\bf Warm} & \multicolumn{1}{|c|}{\bf Reas} & \multicolumn{1}{|c|}{\bf Stab} & \multicolumn{1}{|c|}{\bf Domin} & \multicolumn{1}{|c|}{\bf Live} & \multicolumn{1}{|c|}{\bf Cons} & \multicolumn{1}{|c|}{\bf Soci} & \multicolumn{1}{|c|}{\bf Sens} & \multicolumn{1}{|c|}{\bf Vigil} & \multicolumn{1}{|c|}{\bf Abst}\\ \hline
\bf Parzen & 49.56(2.2) & 60.56(2.0) & 69.67(2.1) & 69.00(2.1) & 76.22(1.7) & 80.67(2.5) & 54.67(2.4) & 52.44(1.0) & 82.11(2.5) & 67.22(4.0)\\ \hline
\bf DTree & 48.00(1.3) & 50.56(2.1) & 52.67(2.1) & 59.22(2.1) & 70.67(1.9) & 72.56(1.7) & 50.11(2.1) & 48.22(2.3) & 67.33(2.1) & 55.67(1.5)\\ \hline
\bf KNN & 50.00(4.0) & 57.89(1.7) & 68.78(1.6) & 68.89(2.7) & 74.33(1.9) & 81.33(1.9) & 51.44(2.5) & 54.00(1.9) & 81.00(2.0) & 66.78(2.3)\\ \hline
\bf NaiveB & 52.56(1.0) & 52.22(1.0) & 67.33(1.0) & 58.11(1.1) & 70.44(1.0) & 77.44(1.2) & 49.22(2.5) & 57.11(1.5) & 74.67(1.0) & 58.78(1.3)\\ \hline
\bf RF & 49.44(2.8) & 56.44(1.2) & 67.67(2.5) & 69.67(2.5) & 76.11(2.0) & 81.44(1.9) & 48.89(2.3) & 49.89(2.3) & 81.67(2.4) & 59.33(2.8)\\ \hline
\multicolumn{11}{|c|}{}\\ \hline
\multicolumn{1}{|c|}{\bf Trait} & \multicolumn{1}{|c|}{\bf Priv} & \multicolumn{1}{|c|}{\bf Appr} & \multicolumn{1}{|c|}{\bf Open} & \multicolumn{1}{|c|}{\bf Reli} & \multicolumn{1}{|c|}{\bf Perf} & \multicolumn{1}{|c|}{\bf Tens} & \multicolumn{1}{|c|}{\bf Adap} & \multicolumn{1}{|c|}{\bf Intro} & \multicolumn{1}{|c|}{\bf Impet} & \multicolumn{1}{|c|}{\bf Cowa}\\ \hline
\bf Parzen & 38.33(1.2) & 48.78(1.1) & 53.89(2.8) & 57.11(2.8) & 60.22(1.1) & 49.67(1.2) & 55.56(1.3) & 67.11(1.5) & 63.22(1.7) & 50.89(1.7)\\ \hline
\bf DTree & 53.67(2.5) & 49.78(2.1) & 51.78(2.1) & 55.67(2.7) & 61.78(2.1) & 52.00(1.4) & 47.67(2.5) & 60.78(1.9) & 46.78(2.1) & 52.00(2.5)\\ \hline
\bf KNN & 46.56(2.5) & 49.00(1.7) & 60.44(1.3) & 60.22(1.7) & 58.11(1.8) & 47.89(1.4) & 54.78(1.0) & 67.33(1.6) & 60.11(1.4) & 57.56(1.5)\\ \hline
\bf NaiveB & 45.56(4.8) & 42.33(2.0) & 49.78(1.1) & 51.33(1.2) & 51.89(1.9) & 42.78(2.0) & 40.44(3.1) & 57.56(1.0) & 57.22(1.5) & 47.89(1.4)\\ \hline
\bf RF & 47.89(1.7) & 52.00(2.5) & 53.78(2.0) & 57.00(3.4) & 60.78(2.6) & 54.67(1.5) & 51.78(3.0) & 68.44(2.9) & 52.78(2.4) & 47.11(2.7)\\ \hline
\multicolumn{11}{|c|}{\bf Female-Appearance feature-Segmented image}\\ \hline
\multicolumn{1}{|c|}{\bf Trait} & \multicolumn{1}{|c|}{\bf Warm} & \multicolumn{1}{|c|}{\bf Reas} & \multicolumn{1}{|c|}{\bf Stab} & \multicolumn{1}{|c|}{\bf Domin} & \multicolumn{1}{|c|}{\bf Live} & \multicolumn{1}{|c|}{\bf Cons} & \multicolumn{1}{|c|}{\bf Soci} & \multicolumn{1}{|c|}{\bf Sens} & \multicolumn{1}{|c|}{\bf Vigil} & \multicolumn{1}{|c|}{\bf Abst}\\ \hline
\bf Parzen & 53.89(1.5) & 64.22(2.1) & 68.56(1.3) & 70.89(1.1) & 75.89(1.7) & 81.00(1.1) & 49.00(1.5) & 57.44(2.1) & 81.11(1.3) & 67.67(1.3)\\ \hline
\bf DTree & 53.00(2.4) & 52.00(2.7) & 60.33(2.9) & 60.56(2.5) & 62.56(4.0) & 68.67(2.9) & 58.22(3.3) & 47.22(3.3) & 67.11(2.5) & 52.11(2.2)\\ \hline
\bf KNN & 53.67(2.3) & 52.78(2.3) & 67.33(1.7) & 69.56(1.3) & 75.00(1.2) & 81.44(1.5) & 47.78(3.8) & 62.11(1.8) & 81.67(1.6) & 67.22(1.5)\\ \hline
\bf NaiveB & 49.67(1.8) & 56.11(1.2) & 61.78(1.8) & 68.78(1.8) & 68.11(1.5) & 74.67(1.1) & 57.67(2.5) & 50.89(2.7) & 79.56(1.0) & 60.67(1.7)\\ \hline
\bf RF & 50.78(1.3) & 50.56(2.1) & 68.67(1.5) & 70.00(2.1) & 76.00(2.5) & 82.00(1.7) & 52.44(2.1) & 55.11(2.3) & 82.00(1.3) & 60.78(1.9)\\ \hline
\multicolumn{11}{|c|}{}\\ \hline
\multicolumn{1}{|c|}{\bf Trait} & \multicolumn{1}{|c|}{\bf Priv} & \multicolumn{1}{|c|}{\bf Appr} & \multicolumn{1}{|c|}{\bf Open} & \multicolumn{1}{|c|}{\bf Reli} & \multicolumn{1}{|c|}{\bf Perf} & \multicolumn{1}{|c|}{\bf Tens} & \multicolumn{1}{|c|}{\bf Adap} & \multicolumn{1}{|c|}{\bf Intro} & \multicolumn{1}{|c|}{\bf Impet} & \multicolumn{1}{|c|}{\bf Cowa}\\ \hline
\bf Parzen & 44.22(1.8) & 51.11(1.3) & 63.56(1.0) & 60.33(1.3) & 57.22(1.1) & 46.67(1.8) & 55.00(2.5) & 68.33(1.0) & 59.67(2.4) & 49.89(2.2)\\ \hline
\bf DTree & 45.33(1.8) & 43.78(3.3) & 56.89(3.3) & 55.67(3.7) & 57.00(3.1) & 62.89(2.8) & 52.22(2.6) & 61.56(1.9) & 55.11(2.2) & 48.44(2.6)\\ \hline
\bf KNN & 46.11(2.1) & 53.44(1.6) & 65.78(1.6) & 57.56(1.6) & 54.44(1.7) & 50.89(1.9) & 51.00(2.6) & 67.89(1.3) & 54.22(1.7) & 47.56(2.0)\\ \hline
\bf NaiveB & 46.11(2.6) & 52.22(1.9) & 56.22(1.4) & 50.33(1.6) & 46.33(2.1) & 54.00(2.7) & 56.22(1.8) & 61.67(1.8) & 50.78(2.4) & 54.44(1.1)\\ \hline
\bf RF & 46.11(2.3) & 49.56(1.5) & 57.33(2.0) & 56.33(2.1) & 56.56(2.1) & 53.56(2.5) & 51.00(1.7) & 66.22(2.1) & 54.00(1.5) & 51.00(2.3)\\ \hline
\end{tabular}
\label{table3}
\end{adjustwidth}
\end{table}

Comparing the accuracy of each personality trait for men and women, we find only the personality traits ``Dominance'', ``Liveliness'' and ``Vigilance'' have significantly higher accuracy for women than men. For the other personality traits, the performance of the women is either slightly higher than the men, or similar to the men. This result is not fully consistent with those derived from the structural feature, where for most of the personality traits, the performance on women is better than the performance on men. However, for the traits ``Privateness'' and ``Apprehension'', the accuracy for men is much better than the accuracy for women, and the differences are bigger than those of the structural feature, which further suggests that the appearance feature contains richer information in predicting these two traits for men.

For measured intelligence, for both men and women, the classification accuracy is only slightly beyond the level of chance, and for some classification rules, the results are even below 50$\%$ for men. Again, we find the accuracy for women is better than the accuracy for men. This difference is particularly evident for the cropped images. Considering the near-chance level predictions on measured intelligence from the structural and appearance features, we can conclude that predicting measured intelligence from facial features is difficult, if not impossible, even though the prediction accuracy for women is slightly better than for men.

Comparing the classification results of the structural and appearance features, we find the appearance feature generally performs slightly better than the structural feature. This suggests that there is a stronger correlation between the appearance feature and the personality traits, which is consistent with the conclusion in~\cite{bib8}. To extract the discriminative facial appearance feature, we cropped and segmented the face image, respectively. Then, we obtained the classification accuracy of the appearance feature on each personality trait for these two types of images. The results in Table~\ref{table3} show that the segmentation process is more suitable for this classification problem than the cropping process.

\noindent\emph{\textbf{Remark 1}}: We also used the detection results from each of the five texture detectors as the appearance feature to conduct the classifications respectively, and the classification results of the five different features with respect to the personality traits and intelligence are shown in \nameref{S1 Fig} and \nameref{S2 Fig}. The general conclusion is that the results of these five different appearance features are similar to those of our holistic appearance feature reported in this section.

\paragraph{A Comparison between the Structural and the Appearance Features} Here, we compare the performance of the structural and the appearance features for all the personality traits and for measured intelligence. For each of the two types of features, we select the classification method that obtained the best results for most of the personality traits for this feature according to Algorithm~\ref{alg:2}. Note that because segmentation processing achieves better results than the cropping procedure in constructing the appearance feature, the best classification method for all the personality traits was chosen based on the segmented images.

The best classification method selection procedure for all the personality traits is shown in Algorithm~\ref{alg:2}:
\begin{algorithm}
\caption{Selecting the best classification method for all the personality traits.}
\begin{algorithmic}[1]
\Require For a given feature type:
\item Compare the accuracy of the five classification methods with respect to a specific personality trait and mark the maximal one among them;
\item For each classification method, count the number of personality traits with the highest classification accuracy;
\item The classification method with the maximum counted number is regarded as the best classification method for the current feature.
\end{algorithmic}
\label{alg:2}
\end{algorithm}

In this work, given a feature type and a classification method, we can determine the optimal dimension of this feature type for this method using Algorithm~\ref{alg:1}. Then, using Algorithm~\ref{alg:2}, we can choose the best method for a given feature type. Table~\ref{table4} lists the classification results for the men (the second column) and the women (the sixth column), where the best method for a given feature type is marked by a checkmark (``$\surd$''). For a given method and a given feature type, the optimal feature dimension is shown in the third and seventh column.
\begin{table}[!ht]
\footnotesize
\newcommand{\tabincell}[2]{\begin{tabular}{@{}#1@{}}#2\end{tabular}}
\begin{adjustwidth}{-2.25in}{0in}
\caption{
{\bf The optimal dimension of the feature vector for the classification and regression methods with the best prediction performance for the three types of features, respectively. The best method for each type of feature is marked by a checkmark (``$\surd$'').}}
\begin{tabular}{|c|c|c|c|c|c|c|c|c|}
\hline
\multicolumn{1}{|c|}{\bf Gender} & \multicolumn{4}{|c|}{\bf Male} & \multicolumn{4}{|c|}{\bf Female}\\ \hline
\tabincell{c}{\bf Feature \\ \bf type} & \tabincell{c}{\bf Classification \\ \bf methods} & \tabincell{c}{\bf Feature \\ \bf dimension} & \tabincell{c}{\bf Regression \\ \bf methods} & \tabincell{c}{\bf Feature \\ \bf dimension} & \tabincell{c}{\bf Classification \\ \bf methods} & \tabincell{c}{\bf Feature \\ \bf dimension} & \tabincell{c}{\bf Regression \\ \bf methods} & \tabincell{c}{\bf Feature \\ \bf dimension}\\ \hline
{} & \multicolumn{1}{|l|}{Parzen ($\surd$)} & 30 & \multicolumn{1}{|l|}{Linear} & 2 & \multicolumn{1}{|l|}{Parzen ($\surd$)} & 30 & \multicolumn{1}{|l|}{Linear} & 2\\ \hline
{} & \multicolumn{1}{|l|}{DTree} & 30 & \multicolumn{1}{|l|}{Ridge} & 2 & \multicolumn{1}{|l|}{DTree} & 15 & \multicolumn{1}{|l|}{Ridge}& 2\\ \hline
\bf Structural & \multicolumn{1}{|l|}{KNN} & 20 & \multicolumn{1}{|l|}{Lasso} & 2 & \multicolumn{1}{|l|}{KNN} & 30 & \multicolumn{1}{|l|}{Lasso} & 2\\ \hline
{} & \multicolumn{1}{|l|}{NaiveB} & 5 & \multicolumn{1}{|l|}{Pinv} & 2 & \multicolumn{1}{|l|}{NavieB} & 30 & \multicolumn{1}{|l|}{Pinv} & 2\\ \hline
{} & \multicolumn{1}{|l|}{RF} & 20 & \multicolumn{1}{|l|}{KNN} & 2 & \multicolumn{1}{|l|}{RF} & 20 & \multicolumn{1}{|l|}{KNN} & 10\\ \hline
{} & {} & {} & \multicolumn{1}{|l|}{SVM ($\surd$)} & 5 & {} & {} & \multicolumn{1}{|l|}{SVM ($\surd$)} & 5\\ \hline
\multicolumn{9}{|c|}{}\\ \hline
{} & \multicolumn{1}{|l|}{Parzen ($\surd$)} & 5 & \multicolumn{1}{|l|}{Linear} & 2 & \multicolumn{1}{|l|}{Parzen ($\surd$)} & 15 & \multicolumn{1}{|l|}{Linear} & 2\\ \hline
{} & \multicolumn{1}{|l|}{DTree} & 20 & \multicolumn{1}{|l|}{Ridge} & 2 & \multicolumn{1}{|l|}{DTree} & 5 & \multicolumn{1}{|l|}{Ridge}& 2\\ \hline
\bf Appearance & \multicolumn{1}{|l|}{KNN} & 10 & \multicolumn{1}{|l|}{Lasso} & 20 & \multicolumn{1}{|l|}{KNN} & 10 & \multicolumn{1}{|l|}{Lasso ($\surd$)} & 15\\ \hline
{} & \multicolumn{1}{|l|}{NaiveB} & 5 & \multicolumn{1}{|l|}{Pinv} & 2 & \multicolumn{1}{|l|}{NavieB} & 5 & \multicolumn{1}{|l|}{Pinv} & 2\\ \hline
{} & \multicolumn{1}{|l|}{RF} & 20 & \multicolumn{1}{|l|}{KNN} & 30 & \multicolumn{1}{|l|}{RF} & 30 & \multicolumn{1}{|l|}{KNN} & 2\\ \hline
{} & {} & {} & \multicolumn{1}{|l|}{SVM ($\surd$)} & 5 & {} & {} & \multicolumn{1}{|l|}{SVM} & 20\\ \hline
\multicolumn{9}{|c|}{}\\ \hline
{} & \multicolumn{1}{|l|}{Parzen ($\surd$)} & 5 & \multicolumn{1}{|l|}{Linear} & 2 & \multicolumn{1}{|l|}{Parzen ($\surd$)} & 20 & \multicolumn{1}{|l|}{Linear} & 2\\ \hline
{} & \multicolumn{1}{|l|}{DTree} & 25 & \multicolumn{1}{|l|}{Ridge} & 2 & \multicolumn{1}{|l|}{DTree} & 5 & \multicolumn{1}{|l|}{Ridge}& 2\\ \hline
\bf Fingerprint & \multicolumn{1}{|l|}{KNN} & 2 & \multicolumn{1}{|l|}{Lasso} & 2 & \multicolumn{1}{|l|}{KNN} & 30 & \multicolumn{1}{|l|}{Lasso} & 2\\ \hline
{} & \multicolumn{1}{|l|}{NaiveB} & 2 & \multicolumn{1}{|l|}{Pinv} & 2 & \multicolumn{1}{|l|}{NavieB} & 2 & \multicolumn{1}{|l|}{Pinv} & 2\\ \hline
{} & \multicolumn{1}{|l|}{RF} & 20 & \multicolumn{1}{|l|}{KNN} & 20 & \multicolumn{1}{|l|}{RF} & 20 & \multicolumn{1}{|l|}{KNN} & 20\\ \hline
{} & {} & {} & \multicolumn{1}{|l|}{SVM ($\surd$)} & 2 & {} & {} & \multicolumn{1}{|l|}{SVM ($\surd$)} & 5\\ \hline
\end{tabular}
\label{table4}
\end{adjustwidth}
\end{table}

\begin{figure}[h]
  \centering
  {\includegraphics[width=14.0cm,height=6.0cm]{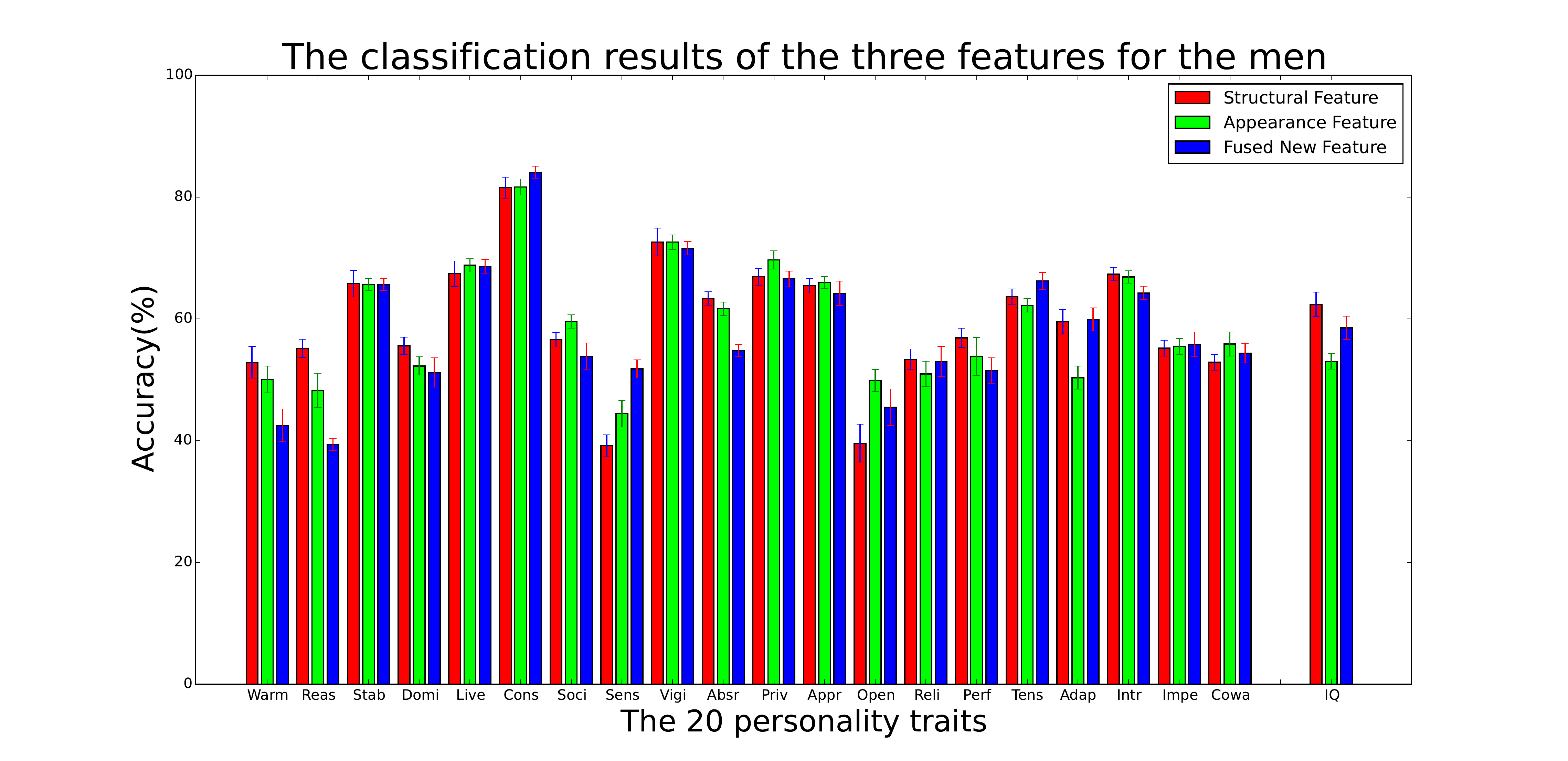}}\\
  {\includegraphics[width=14.0cm,height=6.0cm]{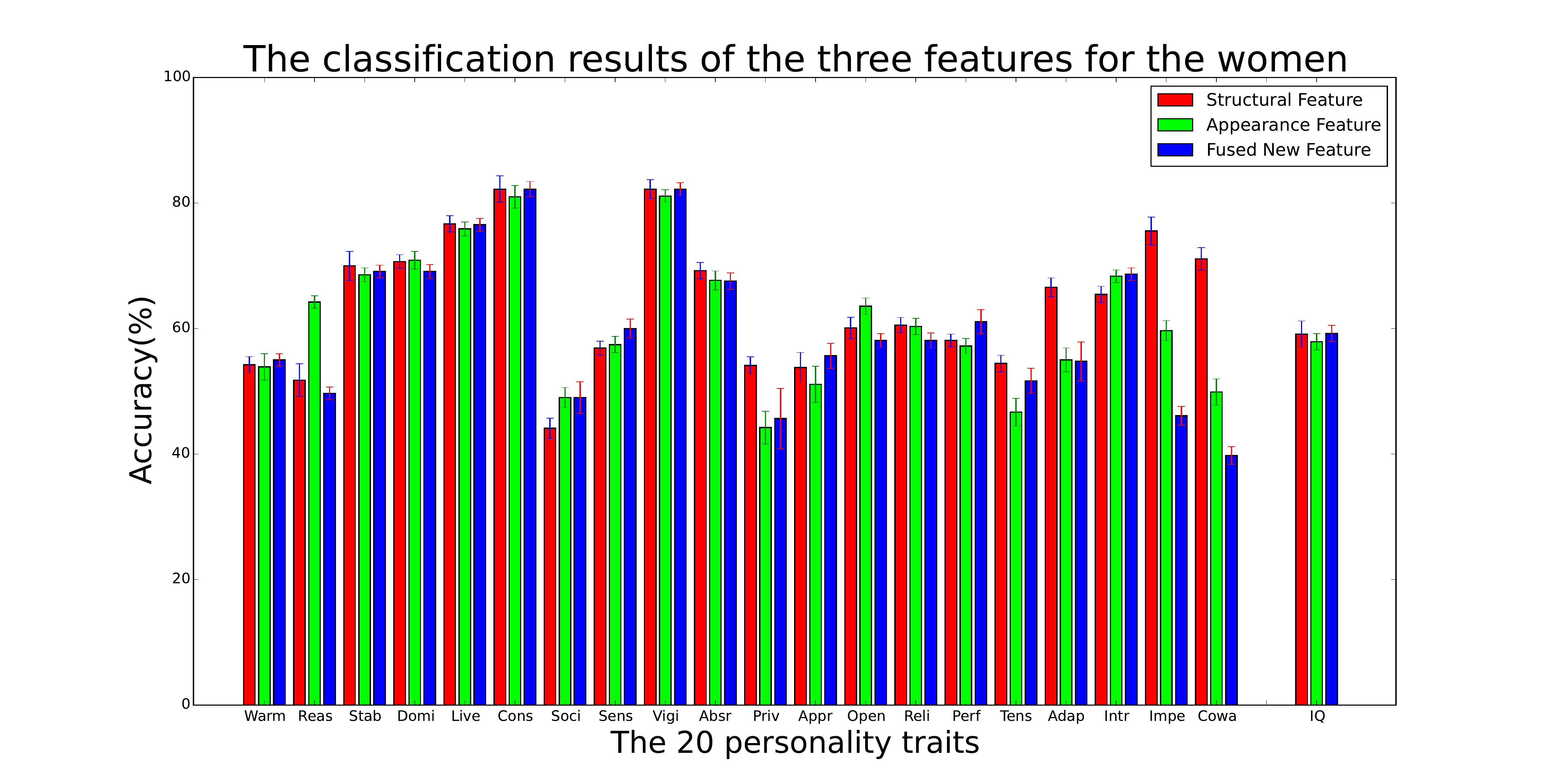}}
\caption{The classification results of the structural feature, the appearance feature and the fused new feature for all the personality traits and for measured intelligence with respect to both genders. The confidence interval is marked by ``I'' at the top of each bar.}
\label{fig:8}
\end{figure}

Fig.~\ref{fig:8} shows the classification results of the above two types of features for all the personality traits and measured intelligence. The performance of these two types of features is generally similar. However, for traits such as ``Adaptation/Anxiety'', the performance of the structural feature is better, while for other traits such as ``Openness'', the performance of the appearance feature is better. Therefore, these two types of features might contain complementary information for predicting the personality traits and intelligence. To test this, each type of feature was reduced to 400 dimensions (mainly determined by the number of samples); then, the 2 reduced features were concatenated into a single new feature, which is itself dimensionally reduced and used for the classification. The classification results of the new feature are shown in Fig.~\ref{fig:8}. Unfortunately, the classification results based on this fused feature are no better than those obtained by the individual features.

\noindent\emph{\textbf{Remark 2}}: In all the above experiments, for a given feature type, we used Algorithm~\ref{alg:1} to select the optimal feature dimension for all the classification methods with respect to all the personality traits. It is understandable that for some personality traits, the selected dimension may not be the best one. Given a personality trait and feature type, different classification methods may have different optimal dimensions. \nameref{S1 Table}, \nameref{S2 Table} and \nameref{S3 Table} are the classification results of personality traits by different methods under different corresponding optimal dimension (marked in parenthesis) for a given feature type. Although some variations exist, the general conclusion of \nameref{S1 Table}, \nameref{S2 Table} and \nameref{S3 Table} are similar to Tables~\ref{table1},~\ref{table2},~\ref{table3} and~\ref{table4}.

\subsection*{The Regression Experiments}
In the regression experiments, because the score of each personality trait measured by the 16PF is a discrete figure ranging from 1 to 10, and the score of the intelligence measured by the Raven's SPM is a percentile value, we directly set these values as the regression targets.

We used six widely used regression methods to investigate whether the self-reported personality traits and measured intelligence can be evaluated from facial features accurately: Linear Regression~\cite{bib38}, Ridge Regression~\cite{bib39}, Lasso Regression~\cite{bib40}, Pseudo-Inverse Regression, K-Nearest Neighbor (KNN) Regression and Support Vector Regression~\cite{bib41}. In our experiments, we used the Radial Basis Function as the kernel of the SVM, and parameter k is set to 5 for the KNN regression following the suggestion in~\cite{bib8}. As for the other parameters of these approaches, they are set empirically by testing many candidate values and choosing the best one for the final experiments. The implementations for these regression methods are off-the-shelf routines from PRTools~\cite{bib37}.

For the regression experiments, becasue our sample number is not large, the regression error is estimated with a 10-fold cross-validation scheme, and the training is repeated thirty times to obtain reliable standard deviations. We adopted several criteria to evaluate the performance of each regression method, including the Root Mean Square Error (RMSE) in Eq.~(\ref{equ:6}) and the Pearson Correlation Coefficient in Eq.~(\ref{equ:7}).

\begin{equation}
RMSE = \sqrt{\frac{\sum_{i=1}^n(X_i-Y_i)^2}{n}}
\label{equ:6}
\end{equation}

\begin{equation}
r = \frac{\sum_{i=1}^n(X_i-\bar{X})(Y_i-\bar Y)}{\sqrt{\sum_{i=1}^n(X_i-\bar{X})^2}\sqrt{\sum_{i=1}^n(Y_i-\bar{Y})^2}}
\label{equ:7}
\end{equation}
where in both~(\ref{equ:6}) and~(\ref{equ:7}), $X_i$ is the self-reported score, $Y_i$ is the predicted score,  $\bar{X}$ and $\bar{Y}$ are the mean scores of the samples, and $n$ is the number of samples.

We also used the residual plots to measure the correlations between the predicted scores and the measured scores of the personality traits and intelligence. When the points representing the residual errors are located randomly around the zero line, it means the predicted scores and the self-reported scores have a significant linear correlation.

\begin{figure}[h]
  \centering
  {\includegraphics[width=6.5cm,height=6.0cm]{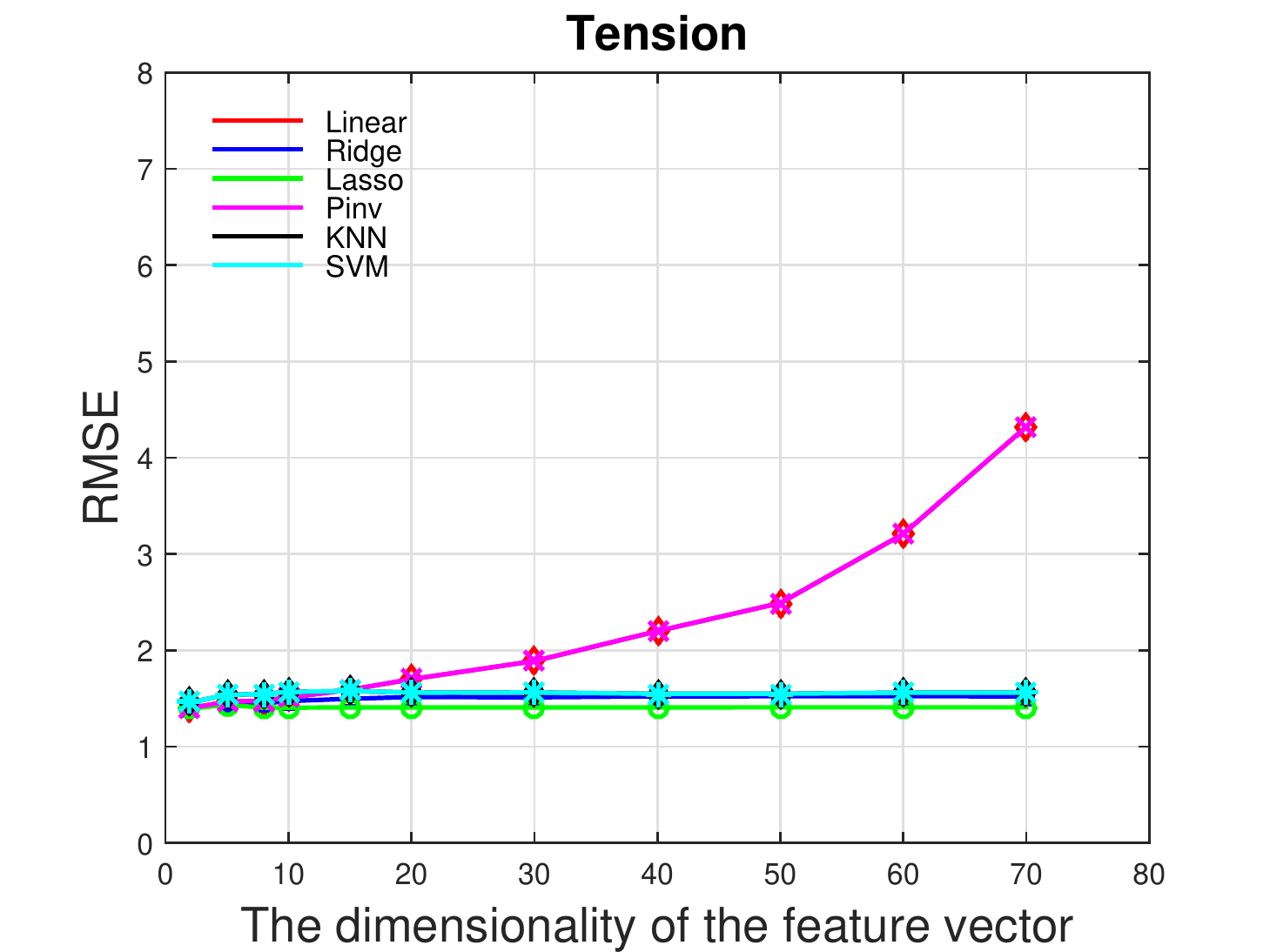}}
  {\includegraphics[width=6.5cm,height=6.0cm]{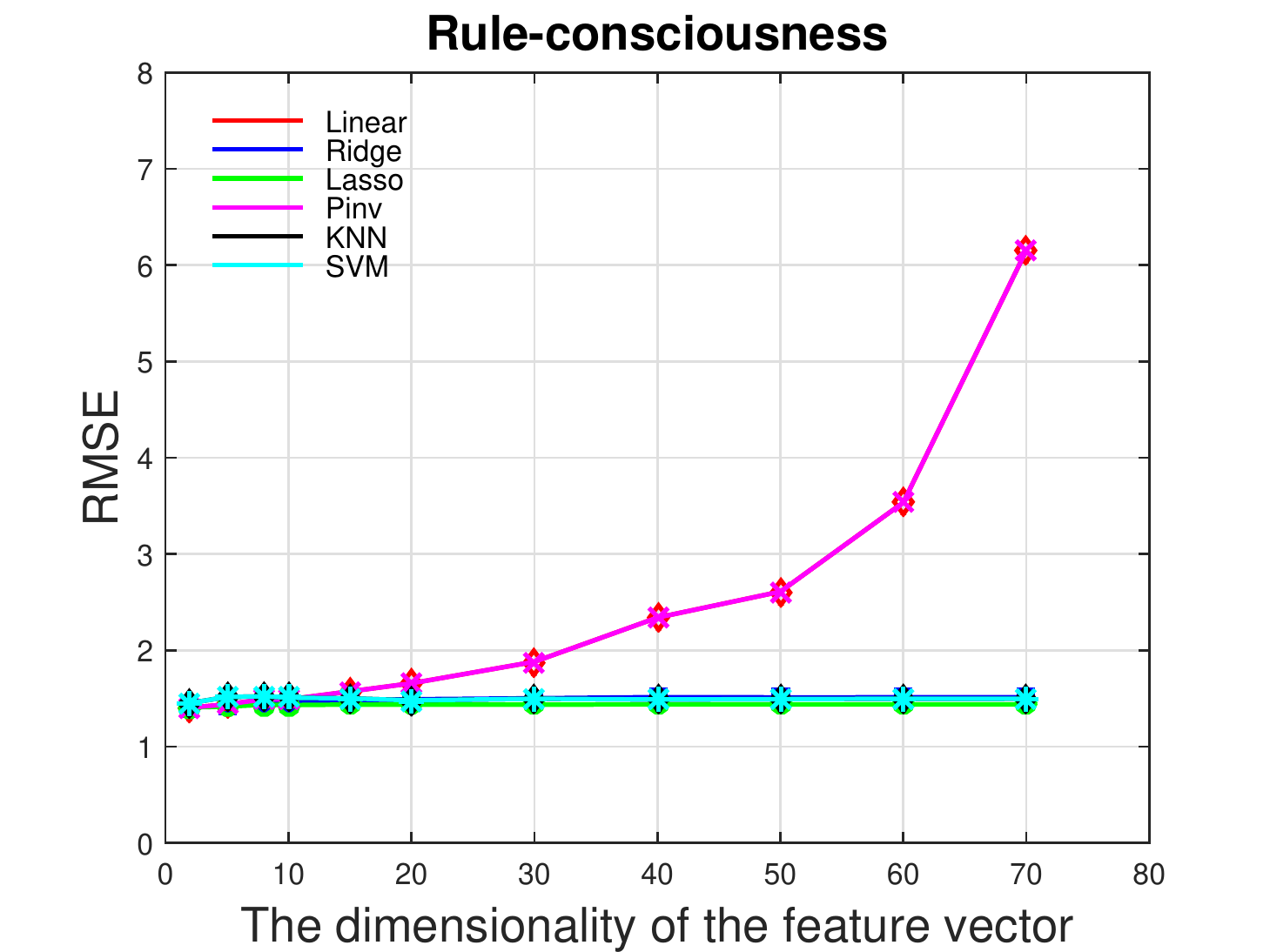}}
\caption{RMSE of the six regression methods on ``Tension'' and ``Rule-consciousness'' under different feature dimensions}
\label{fig:9}
\end{figure}

\paragraph{A Discussion of the Structural Feature} Because the number of available samples was much smaller than the dimension of the structural feature, just as in the classification case, we gradually reduced the dimensions of the original feature to 2, 5, 8, 10, 15, 20, 30, 40, 50, 60 and 70 to empirically select the optimal dimensions using Algorithm~\ref{alg:1} for a given regression method. Optimality was measured by the lowest fitting error. Fig.~\ref{fig:9} shows the RMSE for ``Tension'' and ``Rule-consciousness'' for the six regression methods under different reduced feature dimensions.

Table~\ref{table5} lists the performance of the six regression rules with respect to all the personality traits and to measured intelligence (the corresponding optimal dimension for each regression method shown in Table~\ref{table4}). The results show that ``Rule-consciousness'', ``Sensitivity'' and ``Tension'' have smaller errors than the other personality traits for men. However, ``Openness'', ``Perfectionism'' and ``Tension'' have smaller errors for women. This suggests that due to differences in facial composition between men and women, their structural features have more relation to different personality traits. In addition, for both men and women, the regression errors for ``Social boldness'', ``Vigilance'' and ``Introversion/Extroversion'' are all higher than those of the other traits. This suggests that these three personality traits may have little relationship with the structural feature.
\begin{table}[!ht]
\footnotesize
\begin{adjustwidth}{-2.25in}{0in}
\caption{
{\bf RMSE of the six regression rules for the 20 traits and intelligence with respect to males and females.}}
\begin{tabular}{|l|l|l|l|l|l|l|l|l|l|l|l|}
\hline
\multicolumn{12}{|c|}{\bf Male-Structural feature}\\ \hline
\multicolumn{1}{|c|}{\bf Trait} & \multicolumn{1}{|c|}{\bf Warm} & \multicolumn{1}{|c|}{\bf Reas} & \multicolumn{1}{|c|}{\bf Stab} & \multicolumn{1}{|c|}{\bf Domin} & \multicolumn{1}{|c|}{\bf Live} & \multicolumn{1}{|c|}{\bf Cons} & \multicolumn{1}{|c|}{\bf Soci} & \multicolumn{1}{|c|}{\bf Sens} & \multicolumn{1}{|c|}{\bf Vigil} & \multicolumn{1}{|c|}{\bf Abst} & \multicolumn{1}{|c|}{\bf Intell}\\ \hline
\bf Linear & 1.9683 & 1.5059 & 1.8419 & 1.5304 & 1.8368 & 1.4050 & 1.9373 & 1.4058 & 2.0528 & 1.6105 & 0.2116\\ \hline
\bf Ridge & 1.9412 & 1.4919 & 1.8238 & 1.5236 & 1.8299 & 1.4083 & 1.9692 & 1.3845 & 2.0330 & 1.6055 & 0.2103\\ \hline
\bf Lasso & 1.9411 & 1.4918 & 1.8236 & 1.5235 & 1.8297 & 1.4081 & 1.9690 & 1.3844 & 2.0327 & 1.6053 & 0.2103\\ \hline
\bf Pinv & 1.9683 & 1.5059 & 1.8419 & 1.5304 & 1.8368 & 1.4050 & 1.9373 & 1.4058 & 2.0528 & 1.6105 & 0.2116\\ \hline
\bf KNN & 2.1717 & 1.5705 & 2.0207 & 1.6627 & 1.8760 & 1.4532 & 1.9639 & 1.5255 & 2.0855 & 1.6625 & 0.2271\\ \hline
\bf SVM & 1.9704 & 1.5077 & 1.8102 & 1.5107 & 1.8065 & 1.3932 & 1.9607 & 1.3824 & 2.0011 & 1.5873 & 0.2236\\ \hline
\multicolumn{12}{|c|}{}\\ \hline
\multicolumn{1}{|c|}{\bf Trait} & \multicolumn{1}{|c|}{\bf Priv} & \multicolumn{1}{|c|}{\bf Appr} & \multicolumn{1}{|c|}{\bf Open} & \multicolumn{1}{|c|}{\bf Reli} & \multicolumn{1}{|c|}{\bf Perf} & \multicolumn{1}{|c|}{\bf Tens} & \multicolumn{1}{|c|}{\bf Adap} & \multicolumn{1}{|c|}{\bf Intro} & \multicolumn{1}{|c|}{\bf Impet} & \multicolumn{1}{|c|}{\bf Cowa} & {}\\ \hline
\bf Linear & 1.4441 & 1.7811 & 1.4294 & 1.6628 & 1.4033 & 1.4005 & 1.6576 & 2.0461 & 1.6096 & 1.6110 & {}\\ \hline
\bf Ridge & 1.4286 & 1.7883 & 1.4150 & 1.6550 & 1.4409 & 1.3967 & 1.6811 & 2.0628 & 1.5846 & 1.5943 & {}\\ \hline
\bf Lasso & 1.4285 & 1.7881 & 1.4148 & 1.6548 & 1.4408 & 1.3966 & 1.6809 & 2.0626 & 1.5845 & 1.5942 & {}\\ \hline
\bf Pinv & 1.4441 & 1.7811 & 1.4294 & 1.6628 & 1.4033 & 1.4005 & 1.6576 & 2.0461 & 1.6096 & 1.6110 & {}\\ \hline
\bf KNN & 1.5666 & 1.8167 & 1.5315 & 1.7662 & 1.5233 & 1.4593 & 1.6738 & 2.1074 & 1.7287 & 1.8137 & {}\\ \hline
\bf SVM & 1.4197 & 1.7730 & 1.4023 & 1.6493 & 1.4385 & 1.3897 & 1.6626 & 2.0450 & 1.5948 & 1.5762 & {}\\ \hline
\multicolumn{12}{|c|}{\bf Female-Structural feature}\\ \hline
\multicolumn{1}{|c|}{\bf Trait} & \multicolumn{1}{|c|}{\bf Warm} & \multicolumn{1}{|c|}{\bf Reas} & \multicolumn{1}{|c|}{\bf Stab} & \multicolumn{1}{|c|}{\bf Domin} & \multicolumn{1}{|c|}{\bf Live} & \multicolumn{1}{|c|}{\bf Cons} & \multicolumn{1}{|c|}{\bf Soci} & \multicolumn{1}{|c|}{\bf Sens} & \multicolumn{1}{|c|}{\bf Vigil} & \multicolumn{1}{|c|}{\bf Abst} & \multicolumn{1}{|c|}{\bf Intell}\\ \hline
\bf Linear & 2.0534 & 1.5901 & 1.8515 & 1.6636 & 1.9460 & 1.5113 & 2.1618 & 1.6292 & 2.2623 & 1.6221 & 0.1854\\ \hline
\bf Ridge & 2.0269 & 1.5726 & 1.8319 & 1.6502 & 1.9409 & 1.5052 & 2.1399 & 1.6120 & 2.2429 & 1.6130 & 0.1838\\ \hline
\bf Lasso & 2.0263 & 1.5723 & 1.8314 & 1.6494 & 1.9402 & 1.5049 & 2.1391 & 1.6116 & 2.2423 & 1.6126 & 0.1837\\ \hline
\bf Pinv & 2.0534 & 1.5901 & 1.8515 & 1.6636 & 1.9460 & 1.5113 & 2.1618 & 1.6292 & 2.2623 & 1.6221 & 0.1854\\ \hline
\bf KNN & 2.2021 & 1.6723 & 1.8707 & 1.7566 & 2.0518 & 1.6512 & 2.2533 & 1.7938 & 2.4738 & 1.7497 & 0.1961\\ \hline
\bf SVM & 2.0286 & 1.5685 & 1.8261 & 1.6380 & 1.9648 & 1.4934 & 2.1144 & 1.6046 & 2.2304 & 1.6232 & 0.2062\\ \hline
\multicolumn{12}{|c|}{}\\ \hline
\multicolumn{1}{|c|}{\bf Trait} & \multicolumn{1}{|c|}{\bf Priv} & \multicolumn{1}{|c|}{\bf Appr} & \multicolumn{1}{|c|}{\bf Open} & \multicolumn{1}{|c|}{\bf Reli} & \multicolumn{1}{|c|}{\bf Perf} & \multicolumn{1}{|c|}{\bf Tens} & \multicolumn{1}{|c|}{\bf Adap} & \multicolumn{1}{|c|}{\bf Intro} & \multicolumn{1}{|c|}{\bf Impet} & \multicolumn{1}{|c|}{\bf Cowa} & {}\\ \hline
\bf Linear & 1.6611 & 1.9787 & 1.3263 & 1.5307 & 1.2845 & 1.4736 & 1.8542 & 2.2282 & 1.6534 & 1.5225 & {}\\ \hline
\bf Ridge & 1.6687 & 1.9476 & 1.3260 & 1.5158 & 1.2817 & 1.4547 & 1.8356 & 2.2005 & 1.6352 & 1.5104 & {}\\ \hline
\bf Lasso & 1.6683 & 1.9469 & 1.3257 & 1.5154 & 1.2814 & 1.4540 & 1.8350 & 2.1997 & 1.6347 & 1.5100 & {}\\ \hline
\bf Pinv & 1.6611 & 1.9787 & 1.3263 & 1.5307 & 1.2845 & 1.4736 & 1.8542 & 2.2282 & 1.6534 & 1.5225 & {}\\ \hline
\bf KNN & 1.8246 & 2.0567 & 1.4598 & 1.7150 & 1.3726 & 1.6486 & 1.9588 & 2.3796 & 1.7225 & 1.5613 & {}\\ \hline
\bf SVM & 1.6550 & 1.9420 & 1.3232 & 1.5043 & 1.2977 & 1.4308 & 1.8350 & 2.1990 & 1.6248 & 1.4951 & {}\\ \hline
\end{tabular}
\label{table5}
\end{adjustwidth}
\end{table}

For measured intelligence, the results show that the fitting errors of both genders for all the regression methods are somewhat high, although the errors for women are slightly lower than those for men. This suggests that it is hard to predict a person's intelligence score from the structural feature accurately.

\begin{figure}[h]
\centering
\begin{minipage}[b]{.48\linewidth}
  \centering
  \centerline{\includegraphics[width=7.0cm,height=3.5cm]{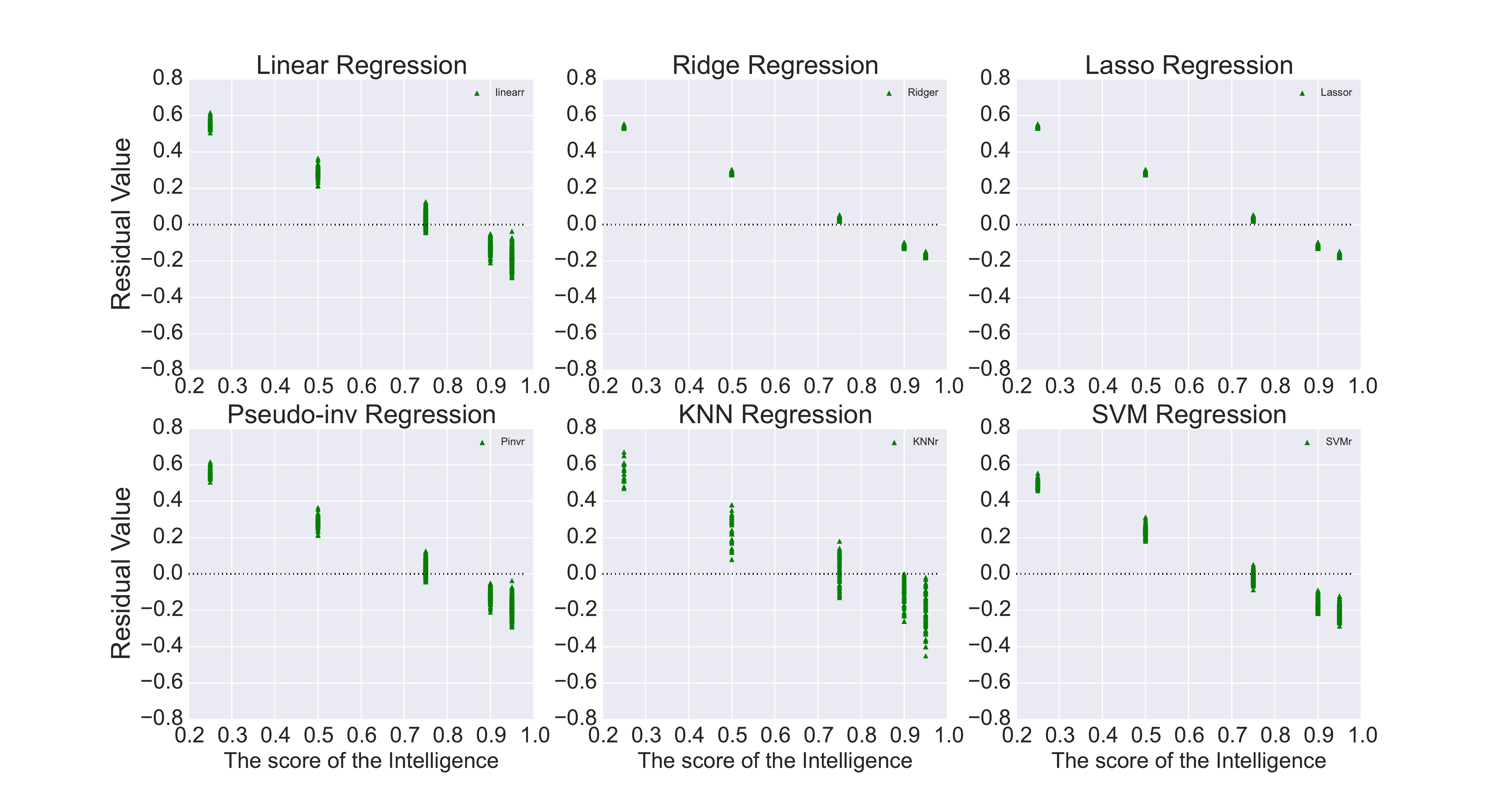}}
  \centerline{(a)}\medskip
\end{minipage}
\hfill
\begin{minipage}[b]{.48\linewidth}
  \centering
  \centerline{\includegraphics[width=7.0cm,height=3.5cm]{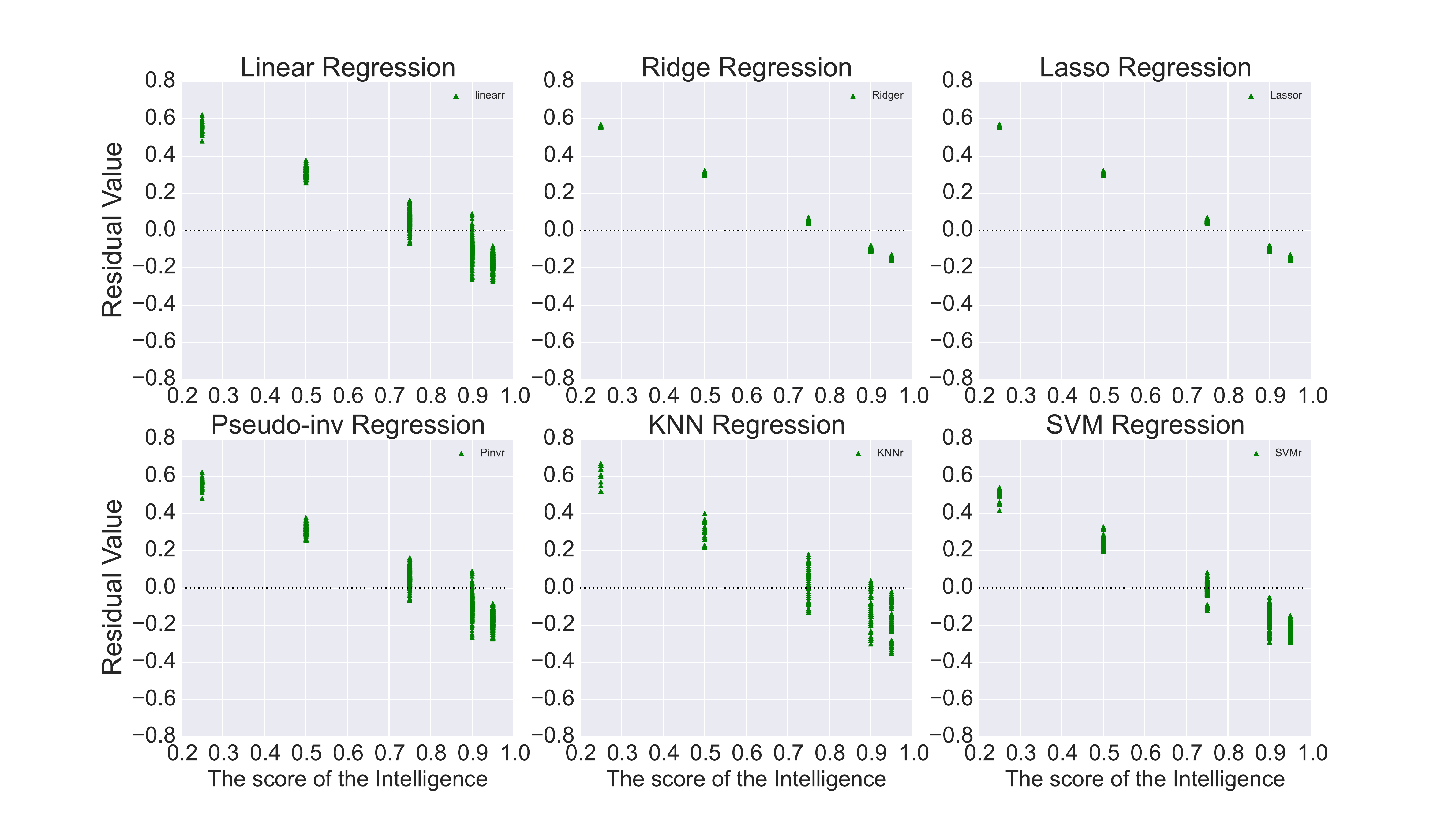}}
  \centerline{(b)}\medskip
\end{minipage}
\\
\begin{minipage}[b]{0.48\linewidth}
  \centering
  \centerline{\includegraphics[width=7.0cm,height=3.5cm]{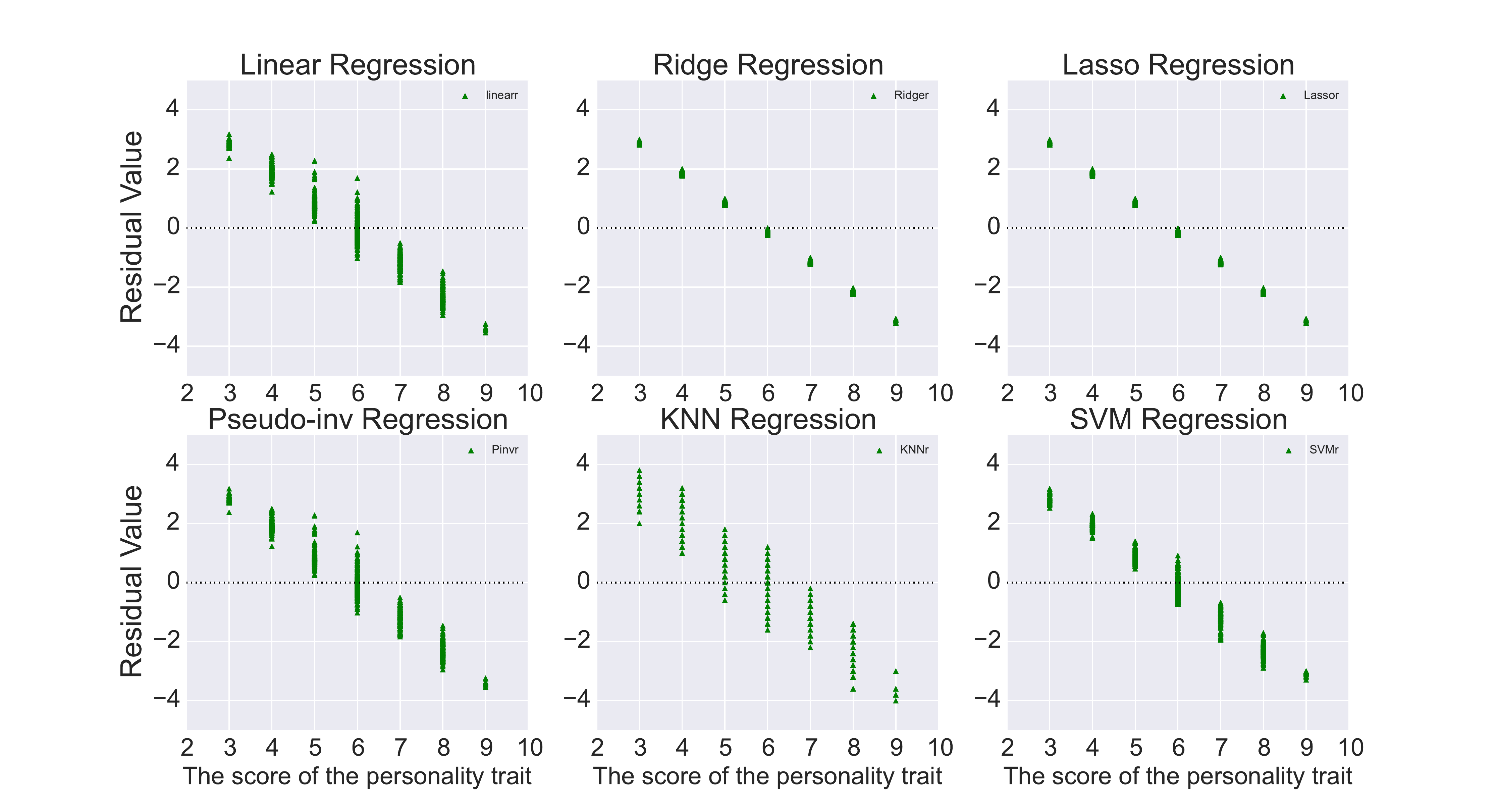}}
  \centerline{(c)}\medskip
\end{minipage}
\hfill
\begin{minipage}[b]{0.48\linewidth}
  \centering
  \centerline{\includegraphics[width=7.0cm,height=3.5cm]{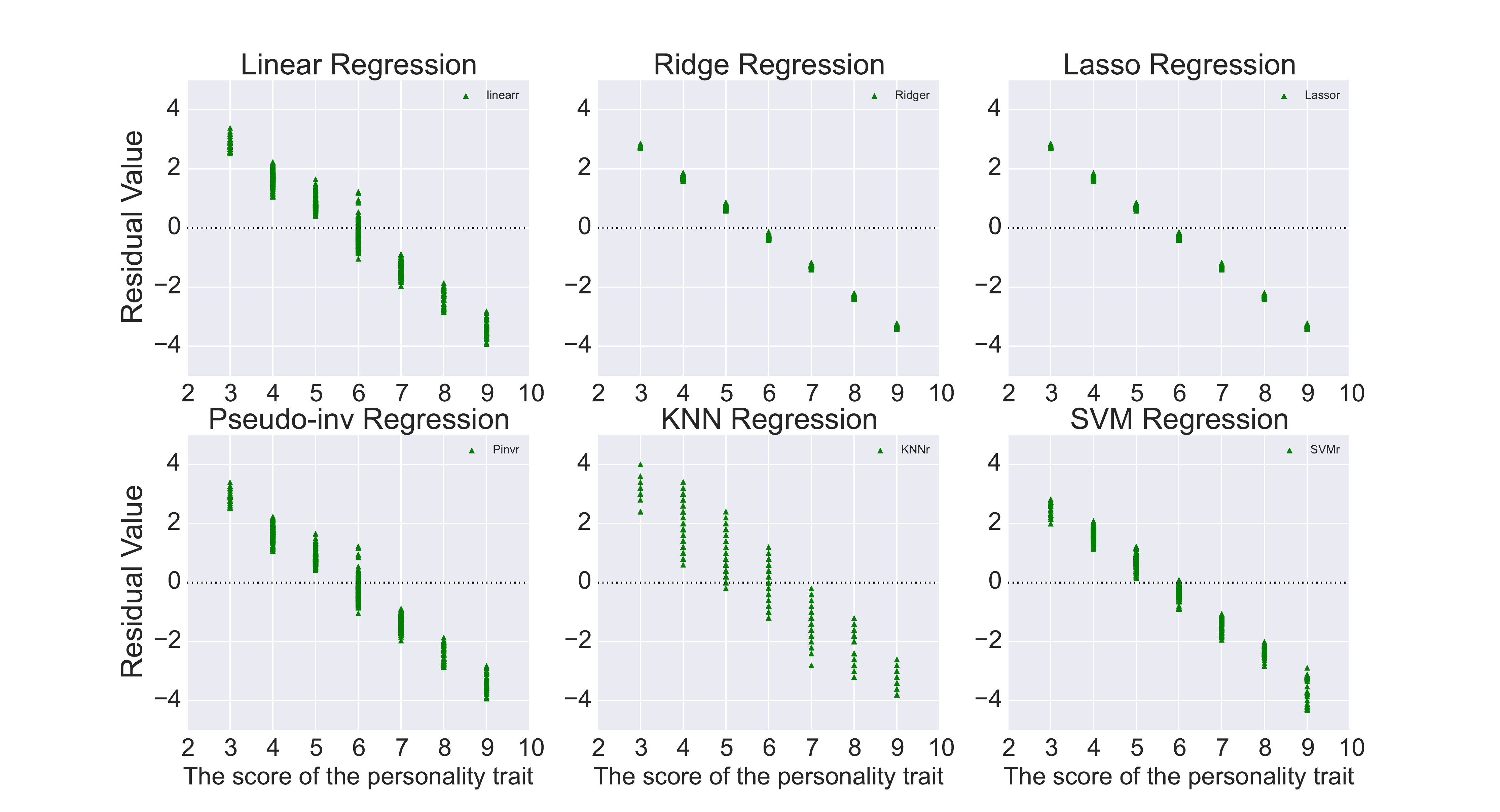}}
  \centerline{(d)}\medskip
\end{minipage}
\caption{The residual plots for the six regresion methods. The images are as follows: (a) the results on ``Intelligence'' for men; (b) the results on ``Intelligence'' for women; (c) the results on ``Tension'' for men; (d) the results on ``Tension'' for women.}
\label{fig:10}
\end{figure}

To observe the correlation between the self-reported scores and the predicted scores of the personality traits and measured intelligence, we draw the residual plots for all the regression approaches. Because our sample number is small, we once again used the cross-validation and repeated-experiment strategies to alleviate the problem. The residual plots on different personality traits are generally similar. As an example, Fig.~\ref{fig:10} shows the residual plots of ``Intelligence'' and ``Tension'' for both men and women, respectively.

Our general conclusion on these residual plots is that the linear correlations between the measured scores and the predicted scores of the personality traits and intelligence are not strong. More specifically, we find that for all the personality traits, the predicted scores of most of the samples are concentrated in several median values. Thus, the regression errors of the median values are small while the errors at both ends are larger. In our work, we also calculate the Pearson Correlation Coefficient between the measured scores and the predicted scores of the personality traits and intelligence, but the correlations are not strong either (r $<$ 0.20, p $<$ 0.01).

For some personality traits, humans do not favor a clear-cut choice in general. For example, people often remark that he or she is a bit sensitive or somewhat domineering. Besides as our samples are not large, and most scores for the personality traits are 4, 5 and 6 (note that this result could stem from the bias that our samples are college students), the regression models trained on our dataset are more suitable for the prediction of the median scores. Because the observational and reasoning ability of undergraduates generally measure higher on intelligence tests (most of the intelligence scores were located in the upper-value intervals), the residual plots also indicate that the prediction performance of the regression models is better for higher scores.

\paragraph{A Discussion of the Appearance Feature} The experiment using the appearance feature was similar to that for the structural feature. The dimensions of the appearance feature were gradually reduced to 2, 5, 8, 10, 15, 20, 30, 40, 50, 60 and 70 using PCA to select the optimal dimensionality for a given regression method by Algorithm~\ref{alg:1}. Then, we measured the regression error for each regression method. Here, we performed the experiments on the cropped and segmented face images separately. Table~\ref{table4} shows the obtained optimal dimension for each one of the six regression methods. Table~\ref{table6} and~\ref{table7} show the results of the six regression methods with respect to each personality trait and intelligence. The conclusions are similar to those obtained for the structural feature. The results show that ``Rule-consciousness'', ``Sensitivity'' and ``Tension'' have smaller fitting errors for men, while ``Openness'', ``Perfectionism'' and ``Tension'' have smaller fitting errors for women, and the fitting errors of ``Social boldness'', ``Vigilance'' and ``Introversion/Extroversion'' are all higher than the other personality traits for both genders.
\begin{table}[!ht]
\footnotesize
\begin{adjustwidth}{-2.25in}{0in}
\caption{
{\bf RMSE of the six regression rules for the 20 traits and intelligence with respect to men with for both cropped and segmented images.}}
\begin{tabular}{|l|l|l|l|l|l|l|l|l|l|l|l|}
\hline
\multicolumn{12}{|c|}{\bf Male-Appearance feature-Cropped image}\\ \hline
\multicolumn{1}{|c|}{\bf Trait} & \multicolumn{1}{|c|}{\bf Warm} & \multicolumn{1}{|c|}{\bf Reas} & \multicolumn{1}{|c|}{\bf Stab} & \multicolumn{1}{|c|}{\bf Domin} & \multicolumn{1}{|c|}{\bf Live} & \multicolumn{1}{|c|}{\bf Cons} & \multicolumn{1}{|c|}{\bf Soci} & \multicolumn{1}{|c|}{\bf Sens} & \multicolumn{1}{|c|}{\bf Vigil} & \multicolumn{1}{|c|}{\bf Abst} & \multicolumn{1}{|c|}{\bf Intell}\\ \hline
\bf Linear & 1.9720 & 1.4870 & 1.8338 & 1.5306 & 1.8423 & 1.4203 & 1.9450 & 1.4124 & 2.0527 & 1.6098 & 0.2113\\ \hline
\bf Ridge & 1.9569 & 1.4835 & 1.8175 & 1.5225 & 1.8197 & 1.4001 & 1.9430 & 1.3913 & 2.0125 & 1.5897 & 0.2113\\ \hline
\bf Lasso & 1.9323 & 1.4686 & 1.7986 & 1.5224 & 1.8080 & 1.4060 & 1.9607 & 1.3827 & 1.9851 & 1.5692 & 0.2086\\ \hline
\bf Pinv & 1.9720 & 1.4870 & 1.8338 & 1.5306 & 1.8423 & 1.4203 & 1.9450 & 1.4124 & 2.0527 & 1.6098 & 0.2113\\ \hline
\bf KNN & 2.2203 & 1.5943 & 1.8184 & 1.4861 & 1.8738 & 1.4628 & 2.0958 & 1.5001 & 2.1509 & 1.7177 & 0.2310\\ \hline
\bf SVM & 1.9721 & 1.5066 & 1.8078 & 1.5134 & 1.8073 & 1.3903 & 1.9568 & 1.3795 & 1.9895 & 1.5793 & 0.2234\\ \hline
\multicolumn{12}{|c|}{}\\ \hline
\multicolumn{1}{|c|}{\bf Trait} & \multicolumn{1}{|c|}{\bf Priv} & \multicolumn{1}{|c|}{\bf Appr} & \multicolumn{1}{|c|}{\bf Open} & \multicolumn{1}{|c|}{\bf Reli} & \multicolumn{1}{|c|}{\bf Perf} & \multicolumn{1}{|c|}{\bf Tens} & \multicolumn{1}{|c|}{\bf Adap} & \multicolumn{1}{|c|}{\bf Intro} & \multicolumn{1}{|c|}{\bf Impet} & \multicolumn{1}{|c|}{\bf Cowa} & {}\\ \hline
\bf Linear & 1.4358 & 1.7777 & 1.4137 & 1.6693 & 1.4342 & 1.4210 & 1.6638 & 2.0376 & 1.6168 & 1.6157 & {}\\ \hline
\bf Ridge & 1.4211 & 1.7668 & 1.4024 & 1.6500 & 1.4387 & 1.3980 & 1.6614 & 2.0290 & 1.5999 & 1.5937 & {}\\ \hline
\bf Lasso & 1.4294 & 1.7661 & 1.3773 & 1.6651 & 1.4373 & 1.4015 & 1.6719 & 2.0489 & 1.5902 & 1.5560 & {}\\ \hline
\bf Pinv & 1.4358 & 1.7777 & 1.4137 & 1.6693 & 1.4342 & 1.4210 & 1.6638 & 2.0376 & 1.6168 & 1.6157 & {}\\ \hline
\bf KNN & 1.5759 & 1.9320 & 1.4653 & 1.7591 & 1.4309 & 1.4263 & 1.6842 & 2.1113 & 1.6730 & 1.7619 & {}\\ \hline
\bf SVM & 1.4231 & 1.7696 & 1.4038 & 1.6549 & 1.4355 & 1.3961 & 1.6632 & 2.0322 & 1.5992 & 1.5796 & {}\\ \hline
\multicolumn{12}{|c|}{\bf Male-Appearance feature-Segmented image}\\ \hline
\multicolumn{1}{|c|}{\bf Trait} & \multicolumn{1}{|c|}{\bf Warm} & \multicolumn{1}{|c|}{\bf Reas} & \multicolumn{1}{|c|}{\bf Stab} & \multicolumn{1}{|c|}{\bf Domin} & \multicolumn{1}{|c|}{\bf Live} & \multicolumn{1}{|c|}{\bf Cons} & \multicolumn{1}{|c|}{\bf Soci} & \multicolumn{1}{|c|}{\bf Sens} & \multicolumn{1}{|c|}{\bf Vigil} & \multicolumn{1}{|c|}{\bf Abst} & \multicolumn{1}{|c|}{\bf Intell}\\ \hline
\bf Linear & 2.0066 & 1.5233 & 1.8399 & 1.5184 & 1.8613 & 1.4258 & 1.9211 & 1.4107 & 2.0525 & 1.6133 & 0.2129\\ \hline
\bf Ridge & 1.9858 & 1.5047 & 1.8335 & 1.5103 & 1.8443 & 1.4229 & 1.9073 & 1.3998 & 2.0275 & 1.5998 & 0.2105\\ \hline
\bf Lasso & 1.9456 & 1.5274 & 1.8334 & 1.4970 & 1.8408 & 1.4340 & 1.9285 & 1.3790 & 2.0174 & 1.5767 & 0.2102\\ \hline
\bf Pinv & 2.0066 & 1.5233 & 1.8399 & 1.5184 & 1.8613 & 1.4258 & 1.9211 & 1.4107 & 2.0525 & 1.6133 & 0.2129\\ \hline
\bf KNN & 2.2533 & 1.6686 & 2.0146 & 1.4626 & 2.0315 & 1.5216 & 1.9593 & 1.4321 & 2.2757 & 1.6744 & 0.2301\\ \hline
\bf SVM & 1.9679 & 1.4996 & 1.8152 & 1.5193 & 1.8112 & 1.3991 & 1.9554 & 1.3845 & 2.0119 & 1.5772 & 0.2232\\ \hline
\multicolumn{12}{|c|}{}\\ \hline
\multicolumn{1}{|c|}{\bf Trait} & \multicolumn{1}{|c|}{\bf Priv} & \multicolumn{1}{|c|}{\bf Appr} & \multicolumn{1}{|c|}{\bf Open} & \multicolumn{1}{|c|}{\bf Reli} & \multicolumn{1}{|c|}{\bf Perf} & \multicolumn{1}{|c|}{\bf Tens} & \multicolumn{1}{|c|}{\bf Adap} & \multicolumn{1}{|c|}{\bf Intro} & \multicolumn{1}{|c|}{\bf Impet} & \multicolumn{1}{|c|}{\bf Cowa} & {}\\ \hline
\bf Linear & 1.4391 & 1.8125 & 1.4250 & 1.6813 & 1.4638 & 1.4131 & 1.6948 & 2.0419 & 1.6262 & 1.6208 & {}\\ \hline
\bf Ridge & 1.4225 & 1.7916 & 1.4173 & 1.6745 & 1.4485 & 1.4016 & 1.6786 & 2.0267 & 1.6146 & 1.6062 & {}\\ \hline
\bf Lasso & 1.4418 & 1.7904 & 1.4037 & 1.6776 & 1.4464 & 1.3992 & 1.6772 & 2.0294 & 1.5578 & 1.5883 & {}\\ \hline
\bf Pinv & 1.4391 & 1.8125 & 1.4250 & 1.6813 & 1.4638 & 1.4131 & 1.6948 & 2.0419 & 1.6262 & 1.6208 & {}\\ \hline
\bf KNN & 1.5218 & 1.8522 & 1.4793 & 1.7392 & 1.3996 & 1.4389 & 1.7217 & 2.1808 & 1.7117 & 1.5860 & {}\\ \hline
\bf SVM & 1.4126 & 1.7702 & 1.3940 & 1.6567 & 1.4425 & 1.3979 & 1.6619 & 2.0475 & 1.5983 & 1.5814 & {}\\ \hline
\end{tabular}
\label{table6}
\end{adjustwidth}
\end{table}

\begin{table}[!ht]
\footnotesize
\begin{adjustwidth}{-2.25in}{0in}
\caption{
{\bf RMSE of the six regression rules for the 20 traits and intelligence with respect to women for both cropped and segmented images.}}
\begin{tabular}{|l|l|l|l|l|l|l|l|l|l|l|l|}
\hline
\multicolumn{12}{|c|}{\bf Female-Appearance feature-Cropped image}\\ \hline
\multicolumn{1}{|c|}{\bf Trait} & \multicolumn{1}{|c|}{\bf Warm} & \multicolumn{1}{|c|}{\bf Reas} & \multicolumn{1}{|c|}{\bf Stab} & \multicolumn{1}{|c|}{\bf Domin} & \multicolumn{1}{|c|}{\bf Live} & \multicolumn{1}{|c|}{\bf Cons} & \multicolumn{1}{|c|}{\bf Soci} & \multicolumn{1}{|c|}{\bf Sens} & \multicolumn{1}{|c|}{\bf Vigil} & \multicolumn{1}{|c|}{\bf Abst} & \multicolumn{1}{|c|}{\bf Intell}\\ \hline
\bf Linear & 2.0367 & 1.5950 & 1.8247 & 1.6594 & 1.9610 & 1.5173 & 2.1657 & 1.6338 & 2.2664 & 1.6485 & 0.1815\\ \hline
\bf Ridge & 2.0207 & 1.6413 & 1.7905 & 1.6645 & 1.8965 & 1.5381 & 2.1151 & 1.6071 & 2.2969 & 1.6518 & 0.1787\\ \hline
\bf Lasso & 1.9813 & 1.6027 & 1.8126 & 1.6165 & 1.9661 & 1.5281 & 2.1049 & 1.5951 & 2.2569 & 1.6209 & 0.1812\\ \hline
\bf Pinv & 2.0367 & 1.5950 & 1.8247 & 1.6594 & 1.9610 & 1.5173 & 2.1657 & 1.6338 & 2.2664 & 1.6485 & 0.1815\\ \hline
\bf KNN & 2.1632 & 1.6616 & 1.8666 & 1.8375 & 2.0627 & 1.6607 & 2.2102 & 1.7029 & 2.3529 & 1.7356 & 0.1805\\ \hline
\bf SVM & 2.0323 & 1.5601 & 1.8229 & 1.6425 & 1.9529 & 1.5052 & 2.1184 & 1.6041 & 2.2388 & 1.6162 & 0.2052\\ \hline
\multicolumn{12}{|c|}{}\\ \hline
\multicolumn{1}{|c|}{\bf Trait} & \multicolumn{1}{|c|}{\bf Priv} & \multicolumn{1}{|c|}{\bf Appr} & \multicolumn{1}{|c|}{\bf Open} & \multicolumn{1}{|c|}{\bf Reli} & \multicolumn{1}{|c|}{\bf Perf} & \multicolumn{1}{|c|}{\bf Tens} & \multicolumn{1}{|c|}{\bf Adap} & \multicolumn{1}{|c|}{\bf Intro} & \multicolumn{1}{|c|}{\bf Impet} & \multicolumn{1}{|c|}{\bf Cowa} & {}\\ \hline
\bf Linear & 1.6882 & 1.9711 & 1.3230 & 1.5162 & 1.3212 & 1.4568 & 1.8520 & 2.2268 & 1.6641 & 1.5271 & {}\\ \hline
\bf Ridge & 1.6857 & 1.9167 & 1.3235 & 1.5265 & 1.2746 & 1.4031 & 1.7401 & 2.1725 & 1.6534 & 1.4837 & {}\\ \hline
\bf Lasso & 1.6367 & 1.9630 & 1.3192 & 1.4967 & 1.2989 & 1.4312 & 1.8209 & 2.2386 & 1.6165 & 1.5110 & {}\\ \hline
\bf Pinv & 1.6882 & 1.9711 & 1.3230 & 1.5162 & 1.3212 & 1.4568 & 1.8520 & 2.2268 & 1.6641 & 1.5271 & {}\\ \hline
\bf KNN & 1.9164 & 2.0497 & 1.4735 & 1.6503 & 1.3863 & 1.4253 & 1.8562 & 2.3043 & 1.6473 & 1.5868 & {}\\ \hline
\bf SVM & 1.6498 & 1.9359 & 1.3155 & 1.5037 & 1.2983 & 1.4243 & 1.8255 & 2.2064 & 1.6222 & 1.4965 & {}\\ \hline
\multicolumn{12}{|c|}{\bf Female-Appearance feature-Segmented image}\\ \hline
\multicolumn{1}{|c|}{\bf Trait} & \multicolumn{1}{|c|}{\bf Warm} & \multicolumn{1}{|c|}{\bf Reas} & \multicolumn{1}{|c|}{\bf Stab} & \multicolumn{1}{|c|}{\bf Domin} & \multicolumn{1}{|c|}{\bf Live} & \multicolumn{1}{|c|}{\bf Cons} & \multicolumn{1}{|c|}{\bf Soci} & \multicolumn{1}{|c|}{\bf Sens} & \multicolumn{1}{|c|}{\bf Vigil} & \multicolumn{1}{|c|}{\bf Abst} & \multicolumn{1}{|c|}{\bf Intell}\\ \hline
\bf Linear & 2.0368 & 1.5933 & 1.8664 & 1.6743 & 1.9835 & 1.5134 & 2.1508 & 1.6259 & 2.2410 & 1.6422 & 0.1868\\ \hline
\bf Ridge & 2.0203 & 1.5753 & 1.8493 & 1.6418 & 1.9606 & 1.5092 & 2.1326 & 1.6096 & 2.2263 & 1.6298 & 0.1845\\ \hline
\bf Lasso & 2.0248 & 1.5705 & 1.7951 & 1.5733 & 1.9548 & 1.5096 & 2.1071 & 1.5271 & 2.2470 & 1.6116 & 0.1840\\ \hline
\bf Pinv & 2.0368 & 1.5933 & 1.8664 & 1.6743 & 1.9835 & 1.5134 & 2.1508 & 1.6259 & 2.2410 & 1.6422 & 0.1868\\ \hline
\bf KNN & 2.2069 & 1.5878 & 1.9960 & 1.8400 & 2.0760 & 1.6724 & 2.3063 & 1.7106 & 2.3762 & 1.8355 & 0.1941\\ \hline
\bf SVM & 2.0143 & 1.5733 & 1.8233 & 1.6360 & 1.9680 & 1.4926 & 2.1109 & 1.6047 & 2.2249 & 1.6203 & 0.2064\\ \hline
\multicolumn{12}{|c|}{}\\ \hline
\multicolumn{1}{|c|}{\bf Trait} & \multicolumn{1}{|c|}{\bf Priv} & \multicolumn{1}{|c|}{\bf Appr} & \multicolumn{1}{|c|}{\bf Open} & \multicolumn{1}{|c|}{\bf Reli} & \multicolumn{1}{|c|}{\bf Perf} & \multicolumn{1}{|c|}{\bf Tens} & \multicolumn{1}{|c|}{\bf Adap} & \multicolumn{1}{|c|}{\bf Intro} & \multicolumn{1}{|c|}{\bf Impet} & \multicolumn{1}{|c|}{\bf Cowa} & {}\\ \hline
\bf Linear & 1.6956 & 1.9848 & 1.3222 & 1.5334 & 1.3003 & 1.4480 & 1.8783 & 2.2469 & 1.6549 & 1.5082 & {}\\ \hline
\bf Ridge & 1.6761 & 1.9628 & 1.3211 & 1.5166 & 1.3017 & 1.4347 & 1.8602 & 2.2114 & 1.6405 & 1.5115 & {}\\ \hline
\bf Lasso & 1.6692 & 1.9413 & 1.3047 & 1.5034 & 1.2397 & 1.4521 & 1.8435 & 2.2009 & 1.6196 & 1.4651 & {}\\ \hline
\bf Pinv & 1.6956 & 1.9848 & 1.3222 & 1.5334 & 1.3003 & 1.4480 & 1.8783 & 2.2469 & 1.6549 & 1.5082 & {}\\ \hline
\bf KNN & 1.8478 & 2.0837 & 1.3003 & 1.6301 & 1.3906 & 1.5088 & 1.9509 & 2.3711 & 1.7610 & 1.6771 & {}\\ \hline
\bf SVM & 1.6515 & 1.9408 & 1.3191 & 1.5001 & 1.2963 & 1.4315 & 1.8469 & 2.1969 & 1.6258 & 1.4966 & {}\\ \hline
\end{tabular}
\label{table7}
\end{adjustwidth}
\end{table}

In addition, it can be seen that the appearance feature performs similarly to the structural feature for these six regression methods. This suggests that the structural feature and the appearance feature do not differ much in terms of regression for predicting the personality traits and measured intelligence. For measured intelligence predicted from the appearance feature, the regression error for women is slightly lower than that for men. By comparing the errors of the experiments for the cropped and segmented face images, we find that segmentation can achieve a better prediction than cropping, which shows once again that image segmentation is more effective than cropping for removing irrelevant information.

For the segmented face images, we drew the residual plots for all the regression approaches with respect to each personality trait and to measured intelligence to observe the correlation between their measured scores and the predicted scores. Generally speaking, the residual errors are not located randomly around the zero line, which suggests that there is no obvious linear correlation between the measured scores and the predicted scores of the personality traits and intelligence; it is particularly hard to predict precise scores for intelligence. In addition, the Pearson Correlation Coefficient between the measured score and the predicted score of personality traits and that of intelligence also agrees with the nonlinear correlation conclusions of the residual plots. However, similar to the results for the structural feature, the fitting errors of the median scores of the appearance feature are relatively small for the personality traits, and the errors when predicting the intelligence on high scores are also smaller than those when predicting the intelligence on low scores.

\noindent\emph{\textbf{Remark 3}}: We also used the detection result from each one of the five texture detectors as an appearance feature to conduct the regression separately. The regression results of the five different features with respect to the personality traits and intelligence are shown in \nameref{S3 Fig} and \nameref{S4 Fig}. The general conclusion is that the results of these five different appearance features are similar to those of our holistic appearance feature in this section.

\paragraph{A Comparison between the Structural and the Appearance Features} Similar to the classification experiments, we also compared the performance of the structural feature and the appearance feature for all the personality traits and intelligence. Similarly, Algorithm~\ref{alg:2} was used to select the best regression method for these two types of features with respect to all the personality traits. Note that for the appearance feature, because segmentation processing achieves better results than cropping, the best regression rule is chosen based on the segmented images. The regression results for men (the fourth column) and women (the eighth column) are shown in Table~\ref{table4}. The best method for a given feature type is denoted by a checkmark (``$\surd$"), and for a given method and a given feature type, the optimal feature dimension is shown in the fifth and the ninth columns.

\begin{figure}[h]
  \centering
  {\includegraphics[width=14.0cm,height=6.0cm]{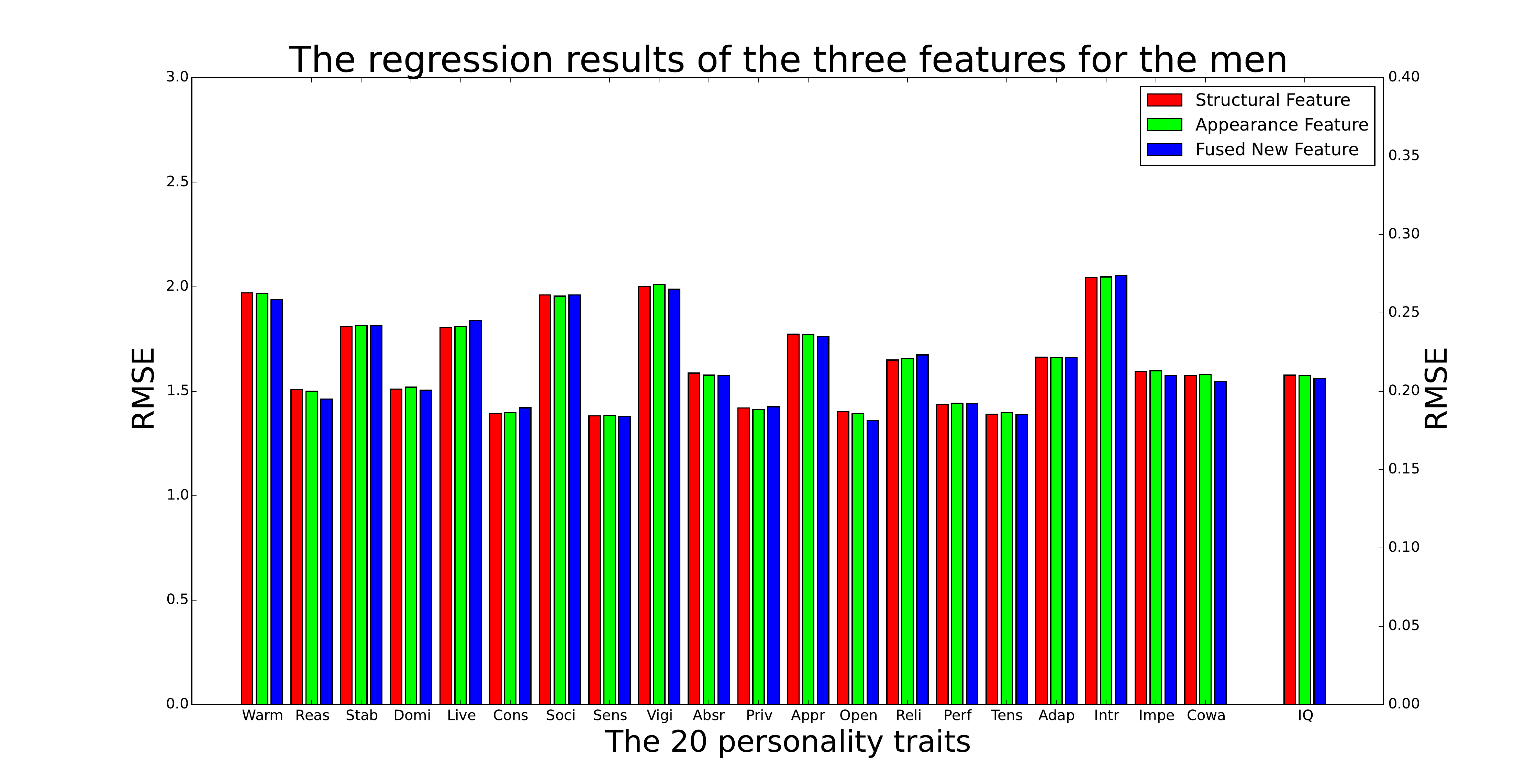}}\\
  {\includegraphics[width=14.0cm,height=6.0cm]{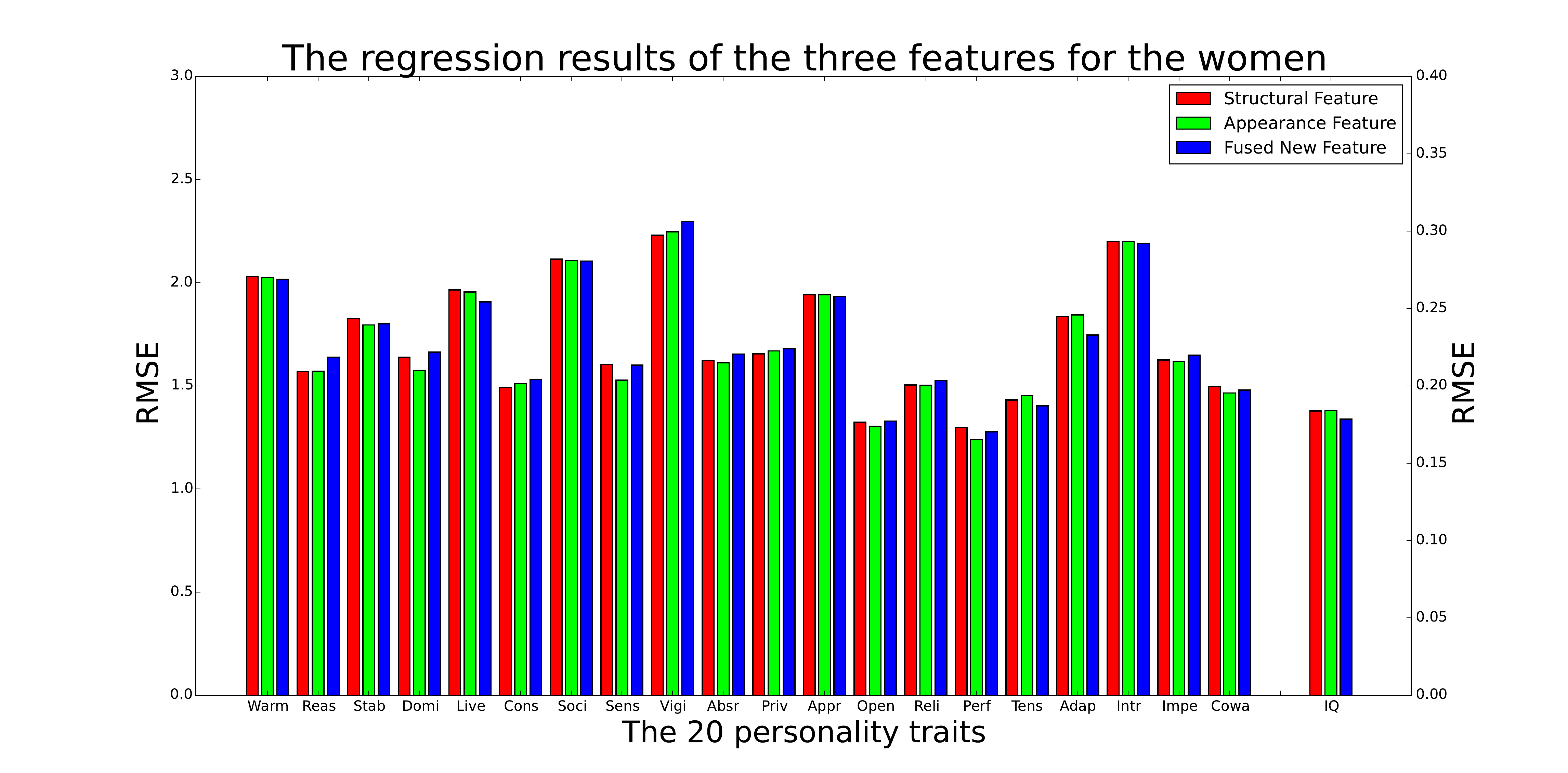}}
\caption{The regression results of the structural feature, the appearance feature and the fused new feature for all the personality traits and for measured intelligence with respect to both genders.}
\label{fig:11}
\end{figure}

Fig.~\ref{fig:11} shows the regression results of these two types of features for all the personality traits and for measured intelligence. In general, they perform comparably. For some traits, such as ``Perfectionism'', the performance of the structural feature is better, while for other traits, such as ``Stability'', the performance of the appearance feature is better. Therefore, these two types of features may contain complementary information for predicting the personality traits and intelligence. As in the classification case and using the same procedure, we constructed a new feature by combining the two types of features. The regression results for this new fused feature are also shown in Fig.~\ref{fig:11}. However, the prediction results using the fused new feature do not reveal any noticeable improvement over the predictions using single features.

\noindent\emph{\textbf{Remark 4}}: As in the classification experiments, given a feature type and a regression method, the feature optimal dimensionality was kept the same for all 20 personality traits in our regression experiments, as shown in Table~\ref{table4}. Clearly, however, different personality traits may have different optimal dimensionality for a given feature type under a given regression method, \nameref{S4 Table}, \nameref{S5 Table} and \nameref{S6 Table} show the results of the six regression rules for these two types of features, respectively. The corresponding optimal feature dimensions are shown in parentheses. The results in \nameref{S4 Table}, \nameref{S5 Table} and \nameref{S6 Table} are generally similar to Tables~\ref{table5},~\ref{table6},~\ref{table7} and~\ref{table4}.

\subsection*{Evaluating Personality Traits by Fingerprint Images}
In this section, we tested the fingerprint images utility in predicting personality traits. As the structural and appearance feature, the fingerprint feature is used for both classification and regression. Table~\ref{table4} shows the obtained optimal dimension for each classification or regression method with respect to all the personality traits. Both the classification and regression results are similar to those of the facial structural and appearance features and are omitted due to space limitations.

Considering the uniqueness of fingerprints and their roots in genetics, personality trait prediction consistency from the fingerprint feature compared with predictions from the structural and appearance features in this work implies, to some extent, that the basis of personality trait evaluations by facial images lies largely in genetics, or is based on those features that are largely determined by nature rather than nurture.

\section*{Conclusion}
In this paper, we investigated whether self-reported personality traits and measured intelligence could be predicted from facial morphometric features or a fingerprint feature. To represent the characteristics of the face images and the fingerprint images, three types of features were extracted. We detected 21 facial salient points using the LBF method to construct the structural feature. After employing image cropping and segmentation techniques to remove irrelevant information, the appearance feature was obtained by fusing the results of five texture descriptors. The fingerprint feature was extracted based on the minutiae contained in the fingerprint images. The evaluation of personality traits was assessed by both classification and regression, in which five classification rules and six regression rules are applied to train the models for the three types of features, respectively. Finally, the classification accuracy and the regression fitness were analyzed.

The results of the classification experiments show that our three extracted types of features are most related to the personality traits ''Rule-consciousness'' and ``Vigilance'', which can be predicted well beyond chance levels for both genders. These two traits may be genetically determined and could be little influenced by the social environment. Generally speaking, the personality traits of females can be predicted more accurately through machine learning methods than male. However, a stronger correlation exists between facial features and the personality traits ``Privateness'' and ``Apprehension'' for men. Our results also suggest that compared with the structural feature, the appearance feature contains more usable information when predicting personality traits. We found that predicting the intelligence level from any of these three types of features is difficult, although women's facial features appear to be more closely related to measured intelligence than men's.

The results of the regression experiments show that predicting exact scores for the personality traits or intelligence is more difficult than classifying them into a binary category. Due to the difference in facial composition between men and women, their facial features are related to different personality traits. In addition, the regression errors of the traits ``Social boldness'', ``Vigilance'' and ``Introversion/Extroversion'' are all too high for all the three types of features, suggesting that these three traits may have little correlation to the facial features. Generally speaking, when predicting the precise scores for the personality traits or for measured intelligence, the three types of features perform similarly. However, similar to the classification experiments, the correlations between intelligence and the features are weak; therefore, it is difficult to predict intelligence scores reliably. The results of the residual plots indicate no evidence for linear correlation between the predicted scores and the measured scores of the personality traits or intelligence, and the Pearson Correlation Coefficients confirm this observed weak correlation. However, we find that for predicting median scores of the personality traits and high scores of intelligence, the fitting performance is relatively better.

In both the classification experiments and regression experiments, the results show that for the appearance feature, the performance on segmented images is better than that on cropped images, which suggests that segmentation processing is more effective at removing irrelevant information.

In this study, we use handcrafted textural descriptors to extract facial features from a single frontal face image and used typical machine learning methods to train the models. However, there are limits in our work in exploring accurate correlations between facial features and personality traits as well as intelligence. For example, a handcrafted feature proven useful for face recognition does not necessarily mean that feature is also more informative for predicting personality traits.In future work, we plan to use the currently widely used deep networks and deep learning to learn suitable features. In addition, we plan to pursue how to exploit 3D face features in our future work.

\section*{Supporting Information}
\paragraph{S1 Fig}\label{S1 Fig} The classification results of the five classification rules based on all the five appearance features for all the personality traits and intelligence with respect to men.

\paragraph{S2 Fig}\label{S2 Fig} The classification results of the five classification rules based on all the five appearance features for all the personality traits and intelligence with respect to women.

\paragraph{S3 Fig}\label{S3 Fig} RMSE of the six regression rules based on all the five appearance features for all the personality traits and intelligence with respect to men.

\paragraph{S4 Fig}\label{S4 Fig} RMSE of the six regression rules based on all the five appearance features for all the personality traits and intelligence with respect to women.

\paragraph{S1 Table}\label{S1 Table} Mean accuracy of the five classification rules based on the structural feature for the 20 traits with respect to men and women.

\paragraph{S2 Table}\label{S2 Table} Mean accuracy of the five classification rules based on the appearance feature for the 20 traits with respect to genders, images cropping and segmenting.

\paragraph{S3 Table}\label{S3 Table} Mean accuracy of the five classification rules based on the two types of features for intelligence with respect to men and women.

\paragraph{S4 Table}\label{S4 Table} RMSE of the six regression rules based on the structural feature for the 20 traits with respect to men and women.

\paragraph{S5 Table}\label{S5 Table} RMSE of the six regression rules based on the appearance feature for the 20 traits with respect to genders, image cropping or segmenting.

\paragraph{S6 Table}\label{S6 Table} RMSE of the six regression rules based on the two types of features for intelligence with respect to men and women.

\section*{Acknowledgments}
We are grateful for all the students of Xiamen University of Technology in China who participated in this study. This work was supported by the Strategic Priority Research Program of the CAS (XDB02070002) and National Natural Science Foundation of China (61421004, 61333015).

\section*{Author Contributions}
Conceived and designed the experiments: ZYH, WG. Performed the experiments: RZQ, WG, HRX. Analyzed the data: RZQ, WG, HRX. Contributed reagents/materials/analysis/tools: RZQ, WG, HRX. Wrote the paper: RZQ, ZYH.

\nolinenumbers

%
%
%

\end{document}